\pdfoutput=1

\documentclass[11pt]{article}

\usepackage[final]{acl}

\usepackage{times}
\usepackage{latexsym}

\usepackage[T1]{fontenc}

\usepackage[utf8]{inputenc}

\usepackage{microtype}

\usepackage{inconsolata}

\usepackage{graphicx}

\usepackage{amsmath}
\usepackage{amsthm}
\usepackage{amsfonts}
\usepackage{enumerate}
\usepackage{xspace}
\usepackage{ifthen}
\usepackage{tabularx}
\usepackage{amssymb}
\usepackage{soul}
\usepackage{booktabs}
\usepackage{multirow}
\usepackage{listings}
\usepackage{xcolor}     
\usepackage{multicol}
\usepackage{tcolorbox}
\usepackage{tcolorbox}
\usepackage{enumitem}
\tcbuselibrary{breakable, skins}

\tcbset{
  promptbox/.style={
    breakable,
    colback=gray!8,
    colframe=gray!50,
    fonttitle=\bfseries,
    title={#1},
    left=6pt, right=6pt, top=4pt, bottom=4pt,
    before upper={\setlength{\parskip}{4pt}\footnotesize},
  }
}
\usepackage{enumitem}
\usepackage{caption}
\usepackage{cuted}   

\title{On the (In)effectiveness of AMR Augmentation for Large Language Models}

\author{Hoa Quynh Nhung Nguyen\thanks{Corresponding author.}\textsuperscript{$\dagger$}, Jacopo Staiano\textsuperscript{$\diamondsuit$}, Michael Sullivan\textsuperscript{$\dagger$} \\
  \textsuperscript{$\dagger$}Saarland University, \textsuperscript{$\diamondsuit$}University of Trento \\
  \texttt{nhungnhq98@gmail.com}, \texttt{jacopo.staiano@unitn.it}, \texttt{msullivan@lst.uni-saarland.de} \\}

\begin{document}
\maketitle
\begin{abstract}

While Abstract Meaning Representation (AMR) has historically improved performance on a range of NLP tasks, the benefit\textemdash or lack thereof\textemdash of AMR augmentation for modern LLMs is thus far unclear.
In this paper, we attempt to reproduce recent work that reported substantial downstream gains from AMR augmentation, finding that these are likely due to specific choices in the experimental settings used: using a consistent and unified protocol for hyperparameter selection, we observe that text-only baselines consistently match or exceed the performance of AMR-augmented models. To investigate this null result, we introduce a perplexity-based probe measuring the degree to which AMR provides an LLM with supplemental relational knowledge not already available to the model. We find that AMR augmentation does not help LLMs improve their understanding of relational content in the sentence, indicating that augmenting these models with AMR offers no clear benefit on downstream tasks.
\end{abstract}

\section{Introduction}
\label{sec_intro}

The Abstract Meaning Representation \citep[AMR;][]{banarescu-etal-2013-abstract} framework is designed to provide structured, interpretable encodings of sentence meaning. AMR makes explicit the individual relations between entities and events that are represented in a given sentence, breaking down complex meaning into more atomic relations. In addition, AMR strips away stylistic and other aspects of plain text that are irrelevant to meaning, exposing only the semantically relevant information.

\begin{figure}[t] 
  \centering
  \includegraphics[width=\columnwidth]{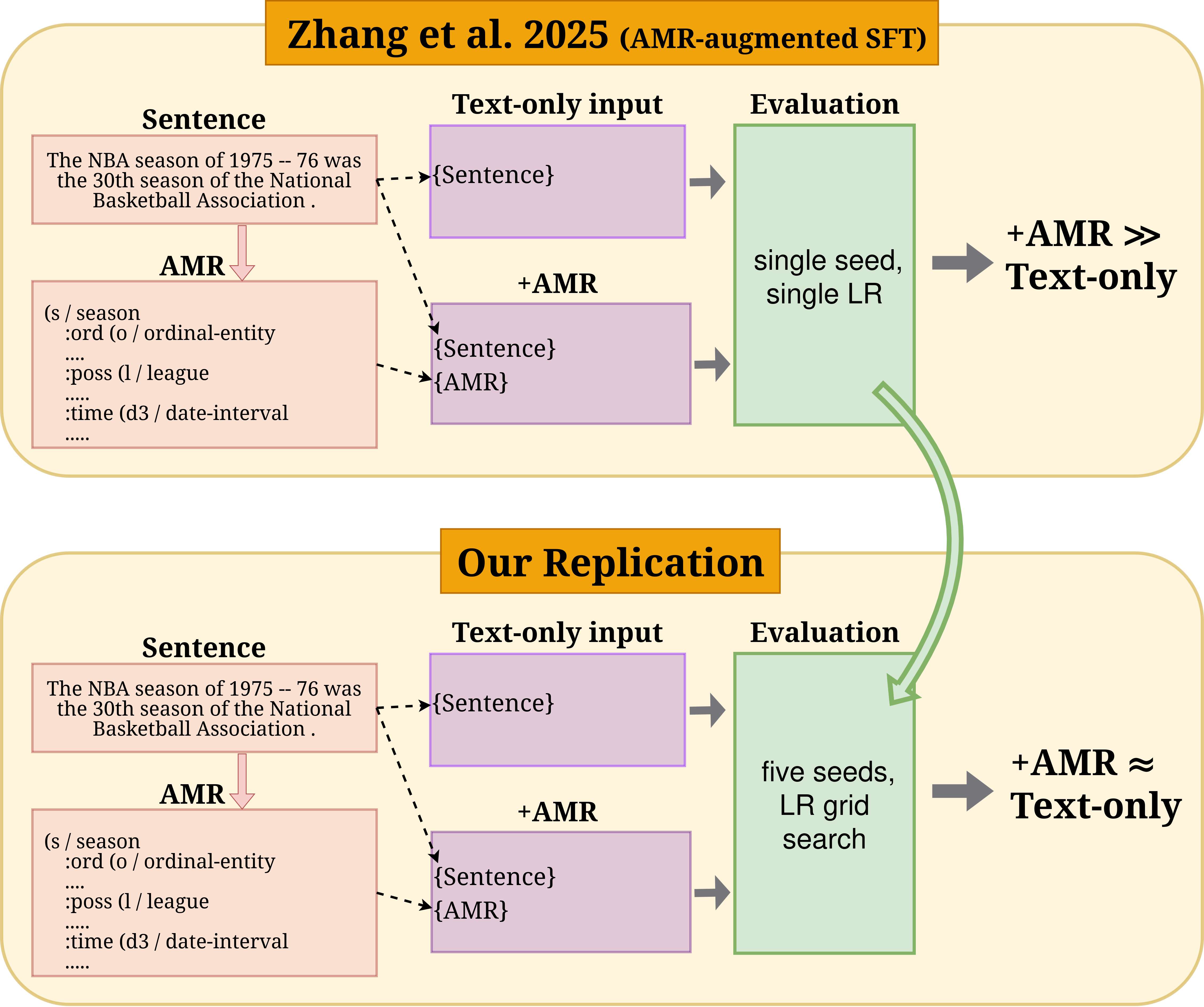} 
  \caption{Prior work (\citeauthor{zhang-etal-2025-sr}, \citeyear{zhang-etal-2025-sr}; top) reports substantial improvement over a text-only baseline from AMR-augmentation. We revisit this method under a more robust evaluation (bottom), finding that AMR augmentation yields no improvement.}
  \label{fig:first_fig}
\end{figure}

This popular semantic representation (SR) format has historically been used to improve performance across various NLP tasks, including Information Retrieval and Extraction \citep{xu-etal-2022-two,zhang-ji-2021-abstract}, Question Answering \citep{bonial-etal-2020-infoforager}, Summarization \citep{hua-etal-2023-improving, hua-etal-2022-amrtvsumm}, and Paraphrase Generation \citep{huang-etal-2022-unsupervised-syntactically}\textemdash in particular in systems built on earlier generations of language models (LMs), such as BERT \citep{devlin-etal-2019-bert} and T5 \citep{raffel2020exploring}.



However, the raw capabilities of modern Large Language Models (LLMs) call into question the benefit of SRs such as AMR for NLP tasks. It is thus far unclear whether SRs can be used to meaningfully augment current, SoTA LLMs: while prior work has shown that AMR representations do not improve the performance of GPT-4 \cite{openai2023gpt} in  a zero-/few-shot setting \cite{jin-etal-2024-analyzing}, \citet{zhang-etal-2025-sr} indicate that AMR augmentation can benefit Llama-3.1-8B-Instruct\footnote{\url{https://huggingface.co/meta-llama/Llama-3.1-8B-Instruct}} \cite{grattafiori2024llama} via supervised fine-tuning (SFT).

In this work, we analyze in-depth the effectiveness of AMR augmentation for open-weight LLMs. First, we find that \textbf{the results reported by \citet{zhang-etal-2025-sr} are not generally reproducible} when reimplementing their experiments (see Figure \ref{fig:first_fig}): the observed performance gains from AMR over the base LLM in that work likely arose from highly specific experimental configurations and custom datasets. In particular, we find no clear improvement from AMR integration, despite substantial fine-tuning using multiple fine-tuning strategies. 

Furthermore, we evaluate AMR integration for more complex, long-text tasks such as multi-sentence event argument extraction, to control for the possibility that the observed failure of AMR integration to improve performance on the simple, single-sentence tasks of \citet{zhang-etal-2025-sr} may be due to saturation on those benchmarks: if the models have already reached peak performance on these tasks, we do not expect any method\textemdash including AMR integration\textemdash to further increase their scores. However, we find that even on these more complex, multi-sentence tasks, AMR integration fails to yield any improvement over the base models.

Given that AMR consistently fails to improve LLM performance across integration strategies and task types, we hypothesize that AMR does not provide an LLM with any relational information that is not already available to the model. To evaluate this hypothesis, we conducted a perplexity-based probe over relational information in AMR graphs. Our findings support our hypothesis, indicating that \textbf{AMR augmentation does not provide any relational information that LLMs are not already capable of extracting} from plain text.



Our contributions are as follows:

\begin{enumerate}
    \item A reimplementation of \citet{zhang-etal-2025-sr}, indicating that the reported results are not generally reproducible and are likely due to specific experimental configurations.
    
    \item An extension of those experiments and results to more complex, multi-sentence tasks.
    
    \item An empirical analysis indicating that AMR augmentation does not provide additional relational knowledge to LLMs beyond what they can obtain from text alone.
\end{enumerate}

We make all code used in these experiments available on GitHub\footnote{\url{https://github.com/nhungachihuuu/AMR-Augmentation-for-LLMs}}.

\section{Related Work} \label{related-work}



\begin{figure}[t] 
  \centering
  \includegraphics[width=\columnwidth]{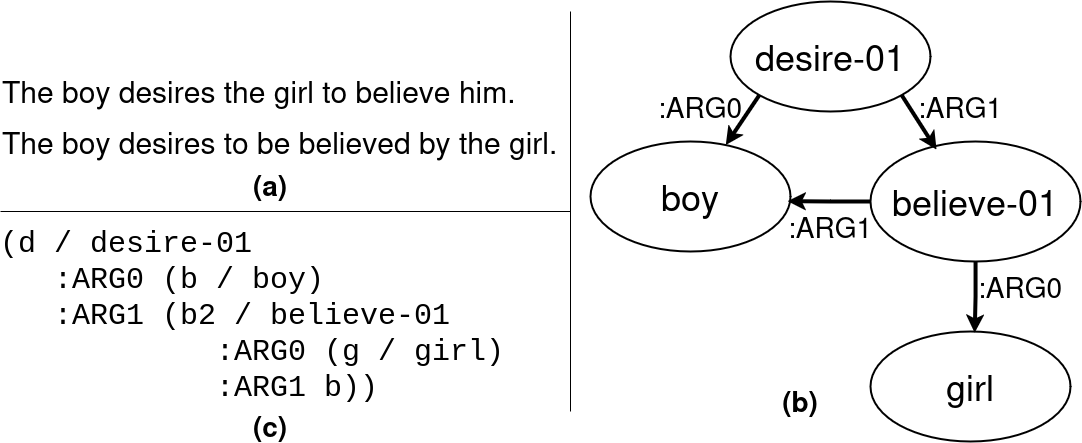} 
  \caption{ Abstract Meaning Representation for two different input sentences (a) with the same semantics. (b) shows the AMR graph and (c) represents the corresponding linearization using PENMAN notation. Example adapted from the AMR Guidelines (\url{https://github.com/amrisi/amr-guidelines}).
 }
  \label{fig:amr}
\end{figure}

\paragraph{Abstract Meaning Representation (AMR).} AMR \citep{banarescu-etal-2013-abstract} is a semantic representation framework that is specifically designed to reflect ``who does what to whom'': the schema centers on 
predicate-argument relations. Using AMR, 
each sentence is represented as a labeled, rooted directed acyclic graph, with nodes 
representing concepts and edges encoding the relations between those concepts (see Figure \ref{fig:amr}). Among graph-based semantic representations such as 
EDS \citep{oepen-lonning-2006-discriminant} and UCCA  \citep{abend-rappoport-2013-universal}, AMR is the most popular and well-resourced 
\citep{wein-opitz-2024-survey}.

Each AMR graph can be converted into a plain-text format using PENMAN notation (\citeauthor{kasper-1989-flexible}, \citeyear{kasper-1989-flexible}; see Figure \ref{fig:amr}): this textual form is referred to as 
\textit{linearized AMR}. The PENMAN format allows AMR to be processed using standard language models, 
which would otherwise require graph-specific architectural modifications.

\paragraph{AMR Applications.} A substantial body of work leveraged AMR to improve model performance on downstream 
NLP tasks for older generations of LMs. The most prevalent application domains are Information Extraction 
\citep{xu-etal-2022-two, zhang-ji-2021-abstract}, Question Answering 
\citep{bonial-etal-2020-infoforager,xu-etal-2021-dynamic}, and Summarization 
\citep{hua-etal-2023-improving, hua-etal-2022-amrtvsumm}. AMR has also been applied to specialized domains such as mathematics \citep{mansouri2022contextualized} and Spatial/Situated Dialogue \citep{bonial-etal-2019-augmenting, bonial-etal-2023-abstract}.

Several studies have explored AMR integration into systems revolving around modern LLMs: \citet{jin-etal-2024-analyzing} incorporated AMR into chain-of-thought 
prompts; \citet{yao-etal-2024-semantic} combined AMR with LLMs for sentence simplification; 
\citet{shi2024compressing} used AMR to enhance retrieval-augmented generation; 
\citet{raut-etal-2025-llms} systematically evaluated LLMs on linearized AMR-incorporated few-shot 
prompts; and \citet{zhang-etal-2025-sr} incorporated linearized AMR into supervised 
fine-tuning.

\paragraph{Analysis of AMR Augmentation.} While the utility of AMR augmentation for older LMs is well-established, the question of its value for modern LLMs remains unanswered.

\citet{jin-etal-2024-analyzing} first investigated the effect of integrating AMR into LLMs: they conducted training-free prompting 
experiments in which the input text is supplemented with its corresponding linearized AMR. Empirical 
results with GPT-3.5 and GPT-4 across multiple tasks show that incorporating AMR into LLMs 
degrades model performance on average, and that it is difficult to predict for a given input whether AMR augmentation will be beneficial.

\citet{zhang-etal-2025-sr}, however, argue that augmenting the LLMs' inputs with linearized AMR\textemdash without fine-tuning the models themselves\textemdash is insufficient to determine the effectiveness of AMR augmentation, as LLMs are not necessarily inherently familiar with this data format. The authors conduct experiments on fine-tuning LLMs with AMR-augmented training data and evaluate their approach on ten tasks\textemdash five of which overlap with 
\citet{jin-etal-2024-analyzing}\textemdash showing that jointly training an LLM on text
augmented with
linearized AMR data yields improvements of up to 12 F1 points. 

Our findings directly contradict those of \citet{zhang-etal-2025-sr}: we are unable to reproduce their results in our reimplementation, despite testing under a broad range of experimental conditions. We instead find evidence extending to fine-tuned settings the findings of \citet{jin-etal-2024-analyzing} that AMR augmentation does not benefit modern LLMs on downstream tasks.

Additionally, to the best of our knowledge, our experiments in Section~\ref{sec_perplexity} represent the first investigation into the reason for the inability of AMR augmentation to improve the downstream performance of LLMs, 
thereby providing additional support to the findings of \citet{jin-etal-2024-analyzing}. 


\section{Single-Sentence Tasks}
\label{sec_1sent}

We first reimplemented the experiments in \citet{zhang-etal-2025-sr}\textemdash simple tasks involving a single\footnote{With the exception of SNLI, PAWS, and WiC, which require classifying \textit{pairs} of single sentences.} input sentence\textemdash in an attempt to reproduce the findings reported in that work. Furthermore, we introduced additional fine-tuning regimens not included in \citet{zhang-etal-2025-sr}, in order to maximize the likelihood of successful AMR augmentation (see e.g. Section \ref{sec_1sent_sub_intermediate}).

\subsection{Methodology}
\label{sec_1sent_sub_method}

Although the code for the experiments in \citet{zhang-etal-2025-sr} is not publicly available, and some experimental settings are under-specified, we mirrored the methodology as closely as possible. Additionally, while \citet{zhang-etal-2025-sr} only conducted experiments with Llama-3.1-8B-Instruct, we included Qwen3-8B\footnote{\url{https://huggingface.co/Qwen/Qwen3-8B}} \cite{yang2025qwen3} in our analysis, to verify the generalizability of the approach to other model families. 

All models were fine-tuned using LoRA \cite{hu2021loralowrankadaptationlarge} with $r = 64$, $\alpha = 128$, and $\text{dropout} = 0.05$, applied to seven target modules: \texttt{q\_proj}, \texttt{k\_proj}, \texttt{v\_proj}, \texttt{o\_proj}, \texttt{gate\_proj}, \texttt{up\_proj}, and \texttt{down\_proj}. Optimal learning rates were selected via grid search for each experimental condition. Additional experimental details are provided in Appendix~\ref{single-sentence-appendix}.

\subsubsection{Datasets}
\label{sec_1sent_sub_method_sub_data}

\citet{zhang-etal-2025-sr} fine-tuned and evaluated on a total of ten datasets: PAWS for paraphrase detection \citep{zhang-etal-2019-paws}, SNLI for entailment 
detection \citep{bowman-etal-2015-large}, WMT16 for machine translation 
\citep{bojar-etal-2016-findings}, CoNLL2003 for named entity recognition 
\citep{tjong-kim-sang-de-meulder-2003-introduction}, SST-2 for sentiment analysis 
\citep{socher-etal-2013-recursive}, PubMed45 for event extraction \citep{garg2016extracting}, 
WiC for word sense disambiguation \citep{pilehvar-camacho-collados-2019-wic}, 
SPIDER for text-to-SQL generation \citep{yu-etal-2018-spider}, AGNews for text 
classification \citep{zhang2015character}, and Logic for logical fallacy 
detection \citep{jin-etal-2022-logical}.

However, the Logic dataset was synthetically generated by \citet{zhang-etal-2025-sr} using GPT-4o-turbo. As the exact prompts used to create this data were not made available, we were unable to reconstruct the Logic dataset, and thus excluded it from our experiments.

For the other nine datasets, we kept the number of training samples as close as possible to those of \citet{zhang-etal-2025-sr}.
For each dataset, we parsed each instance into PENMAN-linearized AMR using \texttt{AMR3-structbart-L} \citep{drozdov-etal-2022-inducing}, a SoTA AMR parser, to ensure high-quality, consistent AMR annotation.

\subsubsection{Fine-Tuning}
\label{sec_1sent_sub_method_sub_fting}

\begin{table}[t]
\centering
\small
\renewcommand{\arraystretch}{1.5}
\scalebox{0.94}{\begin{tabular}{p{1.3cm} p{6cm}}
\toprule
\textbf{Format} & \textbf{Example} \\
\midrule
\multirow{3}{*}{Text-only}
& \textbf{\textit{\{sys\}}} You are an expert in machine translation... \\
& \textbf{\textit{\{user\}}} Obama receives Netanyahu \\
& \textbf{\textit{\{asst\}}} Obama empfängt Netanyahu \\
\midrule
\multirow{5}{*}{+AMR}
& \textbf{\textit{\{sys\}}} You are an expert in machine translation... \\
& \textbf{\textit{\{user\}}} Sentence: Obama receives Netanyahu \\
& 
\textbf{AMR:}\newline 
\texttt{(r / receive-01} \newline
\texttt{\hspace*{1em}:ARG0 (p / person} \newline
\texttt{\hspace*{2em}:name (n / name} \newline
\texttt{\hspace*{3em}:op1 "Obama"))} \newline
\texttt{\hspace*{1em}:ARG1 (p2 / person} \newline
\texttt{\hspace*{2em}:name (n2 / name} \newline
\texttt{\hspace*{3em}:op1 "Netanyahu")))}\\
\bottomrule
\end{tabular}}
\caption{Examples of the text-only (top) and AMR-augmented (bottom) prompt formats for the WMT16 translation task. System messages truncated for illustration; the full prompts are located in Appendix \ref{single_sent_prompts}.}
\label{tab:direct_amr_prompt_examples}
\end{table}

\paragraph{Joint Fine-Tuning.} Following \citet{zhang-etal-2025-sr}, we jointly fine-tuned the models on the training splits of all nine datasets. We fine-tuned baseline models on the original, text-only training data, and fine-tuned AMR-augmented models on a 50:50 mixture of text+AMR and text-only data (see Table \ref{tab:direct_amr_prompt_examples}): this mixture of text-only and AMR-augmented data yielded the best results in \citet{zhang-etal-2025-sr}. At test time, the AMR-tuned models were evaluated only on AMR-augmented examples, as in \citet{zhang-etal-2025-sr}. 

We fine-tuned each model across five seeds, and report the averaged results for evaluation.

\paragraph{Individual Fine-Tuning.} In addition to the joint fine-tuning procedure employed by \citet{zhang-etal-2025-sr}, we further fine-tuned and evaluated baseline (text only) and AMR-augmented models on each dataset individually.

\begin{table*}[h]
\centering
\resizebox{\textwidth}{!}{%
\begin{tabular}{ll|l|ccccccccc}
\toprule
\textbf{Model} & \textbf{Training} & \textbf{Setup} & \textbf{WiC} & \textbf{SST} & \textbf{SNLI} & \textbf{PM} & \textbf{PAWS} & \textbf{AGN} & \textbf{SPD} & \textbf{CNL} & \textbf{WMT} \\
& & & \footnotesize{(F1)} & \footnotesize{(F1)} & \footnotesize{(F1)} & \footnotesize{(F1)} & \footnotesize{(F1)} & \footnotesize{(F1)} & \footnotesize{(EM)} & \footnotesize{(F1)} & \footnotesize{(BLEU)} \\
\midrule
\multirow{6}{*}{Qwen3-8B}
 & \multirow{3}{*}{Joint}      & Text         & $76.8${\tiny$\pm$0.3}          & $\mathbf{96.5}${\tiny$\pm$1.2} & $92.3${\tiny$\pm$0.8}          & $\mathbf{70.4}${\tiny$\pm$2.1} & $92.9${\tiny$\pm$0.7}          & $92.3${\tiny$\pm$0.5}          & $59.2${\tiny$\pm$1.1}          & $\mathbf{92.4}${\tiny$\pm$0.3} & $\mathbf{28.2}${\tiny$\pm$0.5} \\
 &                             & +AMR         & $76.4${\tiny$\pm$0.4}          & $96.1${\tiny$\pm$1.9}          & $92.1${\tiny$\pm$1.2}          & $68.3${\tiny$\pm$2.5}          & $\mathbf{93.0}${\tiny$\pm$0.4} & $91.2${\tiny$\pm$0.3}         & $\mathbf{61.9}${\tiny$\pm$1.2} & $92.1${\tiny$\pm$0.6}          & $\mathbf{28.2}${\tiny$\pm$0.8} \\
 &                             & +AMR +Inter. & $\mathbf{77.5}${\tiny$\pm$1.7} & $96.2${\tiny$\pm$0.3}          & $\mathbf{92.4}${\tiny$\pm$1.3} & $67.2${\tiny$\pm$4.8}          & $92.7${\tiny$\pm$0.7}          & $\mathbf{92.5}${\tiny$\pm$0.3} & $58.8${\tiny$\pm$1.5}          & $92.3${\tiny$\pm$1.6}          & $27.6${\tiny$\pm$0.7}          \\
\cmidrule{2-12}
 & \multirow{3}{*}{Individual} & Text         & $\mathbf{75.9}${\tiny$\pm$2.1} & $96.2${\tiny$\pm$0.5}          & $\mathbf{92.4}${\tiny$\pm$1.2} & $80.2${\tiny$\pm$2.5}          & $\mathbf{94.3}${\tiny$\pm$0.3} & $93.2${\tiny$\pm$0.8}          & $\mathbf{57.8}${\tiny$\pm$1.2} & $93.2${\tiny$\pm$0.5}          & $28.3${\tiny$\pm$0.2}          \\
 &                             & +AMR         & $74.6${\tiny$\pm$1.7}          & $\mathbf{96.8}${\tiny$\pm$1.4} & $91.2${\tiny$\pm$1.3}          & $81.3${\tiny$\pm$1.4}          & $94.1${\tiny$\pm$0.4}          & $\mathbf{93.4}${\tiny$\pm$0.5} & $57.6${\tiny$\pm$1.3}          & $93.3${\tiny$\pm$0.3}          & $\mathbf{28.5}${\tiny$\pm$0.1} \\
 &                             & +AMR +Inter. & $75.1${\tiny$\pm$1.8}          & $96.3${\tiny$\pm$0.2}          & $91.0${\tiny$\pm$1.4}          & $\mathbf{82.9}${\tiny$\pm$1.2} & $\mathbf{94.3}${\tiny$\pm$0.3}          & $93.2${\tiny$\pm$0.2}          & $57.3${\tiny$\pm$1.3}          & $\mathbf{93.4}${\tiny$\pm$0.9} & $28.1${\tiny$\pm$0.1}          \\
\midrule
\multirow{6}{*}{Llama-3.1-8B}
 & \multirow{3}{*}{Joint}      & Text         & $\mathbf{76.7}${\tiny$\pm$1.3} & $96.2${\tiny$\pm$0.8}          & $90.9${\tiny$\pm$1.2}          & $72.1${\tiny$\pm$5.1}          & $\mathbf{93.1}${\tiny$\pm$0.1} & $\mathbf{91.2}${\tiny$\pm$1.1} & $57.1${\tiny$\pm$1.4}          & $92.3${\tiny$\pm$1.6}          & $29.0${\tiny$\pm$0.5}          \\
 &                             & +AMR         & $75.4${\tiny$\pm$2.1}          & $\mathbf{96.5}${\tiny$\pm$0.7} & $\mathbf{91.3}${\tiny$\pm$1.8} & $71.7${\tiny$\pm$7.3}          & $92.6${\tiny$\pm$0.4}          & $90.4${\tiny$\pm$1.2}          & $\mathbf{58.3}${\tiny$\pm$2.8} & $\mathbf{93.8}${\tiny$\pm$0.2} & $\mathbf{29.2}${\tiny$\pm$0.7} \\
 &                             & +AMR +Inter. & $74.9${\tiny$\pm$3.2}          & $96.1${\tiny$\pm$0.9}          & $90.2${\tiny$\pm$1.3}          & $\mathbf{72.4}${\tiny$\pm$7.2} & $92.2${\tiny$\pm$0.8}          & $90.1${\tiny$\pm$1.5}          & $55.2${\tiny$\pm$1.7}          & $92.4${\tiny$\pm$0.7}          & $27.4${\tiny$\pm$0.5}          \\
\cmidrule{2-12}
 & \multirow{3}{*}{Individual} & Text         & $\mathbf{75.4}${\tiny$\pm$2.3} & $\mathbf{96.3}${\tiny$\pm$0.9} & $91.2${\tiny$\pm$0.3}          & $\mathbf{84.1}${\tiny$\pm$0.8} & $\mathbf{93.9}${\tiny$\pm$0.3}          & $\mathbf{93.6}${\tiny$\pm$1.1} & $\mathbf{56.8}${\tiny$\pm$2.5} & $93.2${\tiny$\pm$0.2}          & $\mathbf{29.8}${\tiny$\pm$0.3} \\
 &                             & +AMR         & $75.2${\tiny$\pm$2.1}          & $96.1${\tiny$\pm$0.6}          & $\mathbf{91.3}${\tiny$\pm$0.5} & $82.4${\tiny$\pm$3.2}          & $93.5${\tiny$\pm$0.5} & $93.2${\tiny$\pm$1.2}          & $56.1${\tiny$\pm$1.3}          & $\mathbf{93.7}${\tiny$\pm$1.2} & $29.6${\tiny$\pm$0.3}          \\
 &                             & +AMR +Inter. & $73.8${\tiny$\pm$3.1}          & $95.3${\tiny$\pm$1.2}          & $90.1${\tiny$\pm$1.4}          & $82.6${\tiny$\pm$1.6}          & $93.6${\tiny$\pm$0.3}          & $93.3${\tiny$\pm$0.2}          & $54.5${\tiny$\pm$1.8}          & $93.1${\tiny$\pm$0.9}          & $29.3${\tiny$\pm$0.2}          \\
\midrule
 \citet{zhang-etal-2025-sr} & \multirow{2}{*}{Joint} & Text   & $67.0$\phantom{\tiny$\pm$0.0} & $75.6$\phantom{\tiny$\pm$0.0} & $35.5$\phantom{\tiny$\pm$0.0} & $78.9$\phantom{\tiny$\pm$0.0} & $68.9$\phantom{\tiny$\pm$0.0} & $76.5$\phantom{\tiny$\pm$0.0} & $41.2$\phantom{\tiny$\pm$0.0} & $75.8$\phantom{\tiny$\pm$0.0} & $29.1$\phantom{\tiny$\pm$0.0} \\

 (Llama-3.1-8B) & & +SR & $\mathbf{74.7}$\phantom{\tiny$\pm$0.0} & $\mathbf{83.7}$\phantom{\tiny$\pm$0.0} & $\mathbf{54.9}$\phantom{\tiny$\pm$0.0} & $\mathbf{81.9}$\phantom{\tiny$\pm$0.0} & $\mathbf{81.0}$\phantom{\tiny$\pm$0.0} & $\mathbf{82.6}$\phantom{\tiny$\pm$0.0} & $\mathbf{48.9}$\phantom{\tiny$\pm$0.0} & $\mathbf{76.7}$\phantom{\tiny$\pm$0.0} & $\mathbf{30.3}$\phantom{\tiny$\pm$0.0} \\
\bottomrule
\end{tabular}}%
\caption{Joint and individual training results on the nine single-sentence task datasets for Llama-3.1-8B-Instruct and Qwen3-8B, compared to the results reported in \citet{zhang-etal-2025-sr} for Llama-3.1-8B. The +SR scores from \citet{zhang-etal-2025-sr} are averaged over multiple SR frameworks, including AMR (see the discussion in Section~\ref{sec_1sent_sub_results}). Each cell reports mean $\pm$ standard deviation over runs with different random seeds. The best-performing Setup within each Model, Training type, and dataset is indicated in \textbf{bold}. PM=PubMed45, AGN=AGNews, SPD=SPIDER, CNL=CoNLL2003, WMT=WMT16.}
\label{tab:single_sent_downstream_results}
\end{table*}

\begin{table}[t!]
\centering
\small
\begin{tabular}{lccc}
\toprule
\textbf{Dataset} & \textbf{Llama} & \textbf{Qwen} & \textbf{Pooled} \\
\midrule
\multicolumn{4}{l}{\textit{Individual training}} \\
WMT     & 0.370 & 0.067 & 0.985 \\
PAWS    & 0.166 & 0.273 & 0.064 \\
WiC     & 0.954 & 0.631 & 0.717 \\
SPIDER  & 0.790 & 0.614 & 0.901 \\
AGNews  & 0.590 & 0.297 & 0.211 \\
CoNLL   & 0.433 & 0.110 & 0.750 \\
SNLI    & 0.489 & \textbf{0.033} & 0.234 \\
SST     & 0.999 & 0.789 & 0.773 \\
PubMed45 & 0.244 & 0.206 & 0.787 \\
\midrule
\multicolumn{4}{l}{\textit{Joint training}} \\
WMT     & 0.644 & 0.842 & 0.718 \\
PAWS    & 0.180 & 0.844 & 0.813 \\
WiC     & 0.615 & 0.226 & 0.904 \\
SPIDER  & \textbf{0.017} & 0.191 & 0.199 \\
AGNews  & 0.279 & 0.180 & 0.150 \\
CoNLL   & 0.064 & 0.256 & 0.262 \\
SNLI    & 0.987 & 0.866 & 0.952 \\
SST     & 0.491 & 0.136 & 0.550 \\
PubMed45 & 0.190 & 0.350 & 0.504 \\
\bottomrule
\end{tabular}
\caption{Results ($p$-values) for two-sided $t$-tests comparing the text-only and AMR-augmented (``+AMR'') scores in Table \ref{tab:single_sent_downstream_results}\textemdash significant values ($p<0.05$) are indicated in \textbf{bold}. The ``Pooled'' column denotes a $t$-test over the combined Llama and Qwen results for each task.}
\label{tab:significance}
\end{table}

\paragraph{AMR-to-Text Intermediate Objective.}
\label{sec_1sent_sub_intermediate}
To control for the possibility that the AMR-augmented models may fail to leverage the AMR structures present in their fine-tuning data\textemdash i.e. they may simply ignore the AMR\textemdash we additionally implemented an intermediate fine-tuning procedure to familiarize the models with this representational format. 

Specifically, we first fine-tuned Llama-3.1-8B-Instruct and Qwen3-8B models on AMR-to-text generation, using the AMR 3.0 corpus \citep{amr3} for training and evaluation (full training details and evaluation results for AMR-to-text are provided in Appendix \ref{amr2text}). We then performed the AMR-augmented joint and individual fine-tuning, starting from the checkpoints trained on the intermediate AMR-to-text objective. 

\subsection{Results} 
\label{sec_1sent_sub_results}

Table~\ref{tab:single_sent_downstream_results} presents the joint and individual fine-tuning results for the text-only, AMR-augmented, and intermediate AMR-to-text (see Section~\ref{sec_1sent_sub_intermediate}) Llama-3.1-8B-Instruct and Qwen3-8B models. The joint fine-tuned, AMR-augmented Llama-3.1 model (Llama-3.1, Joint, +AMR in Table~\ref{tab:single_sent_downstream_results}) is a direct reimplementation of the configuration used in \citet{zhang-etal-2025-sr}. We additionally include the results reported by \citet{zhang-etal-2025-sr} for reference. Note that their +SR scores are averaged over three separate SR frameworks, including AMR, as \citet{zhang-etal-2025-sr} do not report a full performance breakdown by SR type. However, \citet{zhang-etal-2025-sr} report that AMR yields the greatest performance gains out of the SRs in their experiments (see their Table 11): if anything, the averaged figures in Table~\ref{tab:single_sent_downstream_results} understate the +AMR performance recorded by those authors.


In contrast to \citet{zhang-etal-2025-sr}, we find no consistent improvement from AMR augmentation across models, tasks, or training settings. Under a two-sided $t$-test, we only find significant ($p<0.05$) differences between text-only and AMR-augmented performance on two of 36 configurations (see Table \ref{tab:significance}): Qwen3-8B under the individual training regime on SNLI (in favor of text-only) and for Llama-3.1-8B under the joint regime on SPIDER (in favor of +AMR). 

The observed discrepancy between our results and those of \citet{zhang-etal-2025-sr} is \textit{not} due to a drastic decrease in our AMR-augmented models' performance relative to the AMR-augmented results reported by those authors: in most cases, our AMR-augmented models actually perform \textit{better} than in the experiments of \citet{zhang-etal-2025-sr}.

Rather, the performance of the baseline, text-only models in \citet{zhang-etal-2025-sr} is substantially lower than the performance of our text-only models. As an example, \citet{zhang-etal-2025-sr} report an F1 score of 35.5 on SNLI for their text-only, fine-tuned Llama-3.1-8B-Instruct model\textemdash this is far lower than we would expect given that even BERT \cite{devlin-etal-2019-bert} reaches $\sim$91\% accuracy on SNLI \cite{nie-etal-2020-adversarial}.

We therefore conclude that the results reported in \citet{zhang-etal-2025-sr} are likely due to a faulty experimental configuration:
the performance of text-only and AMR-augmented LLMs is roughly equal
regardless of the choice of fine-tuning regimen.



\section{Multi-Sentence Tasks}
\label{sec_multisent}

The baseline models' performance on many of the tasks in Section \ref{sec_1sent} is near saturation: on SST-2, SNLI, PAWS, AGNews, and CoNLL2003, the text-only Llama-3.1 and Qwen3 models already achieve accuracy/F1 scores over 90\%. It is therefore possible that AMR augmentation is simply unable to improve model performance on these tasks because there is no more room for improvement.

To account for this possibility, we extended our evaluation of AMR-augmented models to more difficult, multi-sentence tasks\textemdash such as passage-level summarization\textemdash where inputs involve richer and more complex linguistic structures. 

\subsection{Methodology}

\paragraph{Datasets.} We selected four multi-sentence datasets that require understanding complex linguistic and semantic structures: RAMS for event argument extraction \citep{ebner-etal-2020-multi}, CNN/DailyMail for summarization \citep{see2017get}, ANLI for adversarial natural language inference \citep{nie-etal-2020-adversarial}, and LogiQA for reading comprehension \citep{liu2020logiqa}. Dataset statistics are provided in Table \ref{tab:multisentence-datasets}.

As discussed in Section \ref{related-work}, prior work has demonstrated notable improvement from AMR augmentation for earlier generations of LMs on event argument extraction and summarization in particular \citep{xu-etal-2022-two, hua-etal-2023-improving}. If modern LLMs can in fact benefit from AMR augmentation, it therefore stands to reason that we would expect to see the most substantial benefits on such tasks.

\begin{figure}[t] 
  \centering
  \includegraphics[width=\columnwidth]{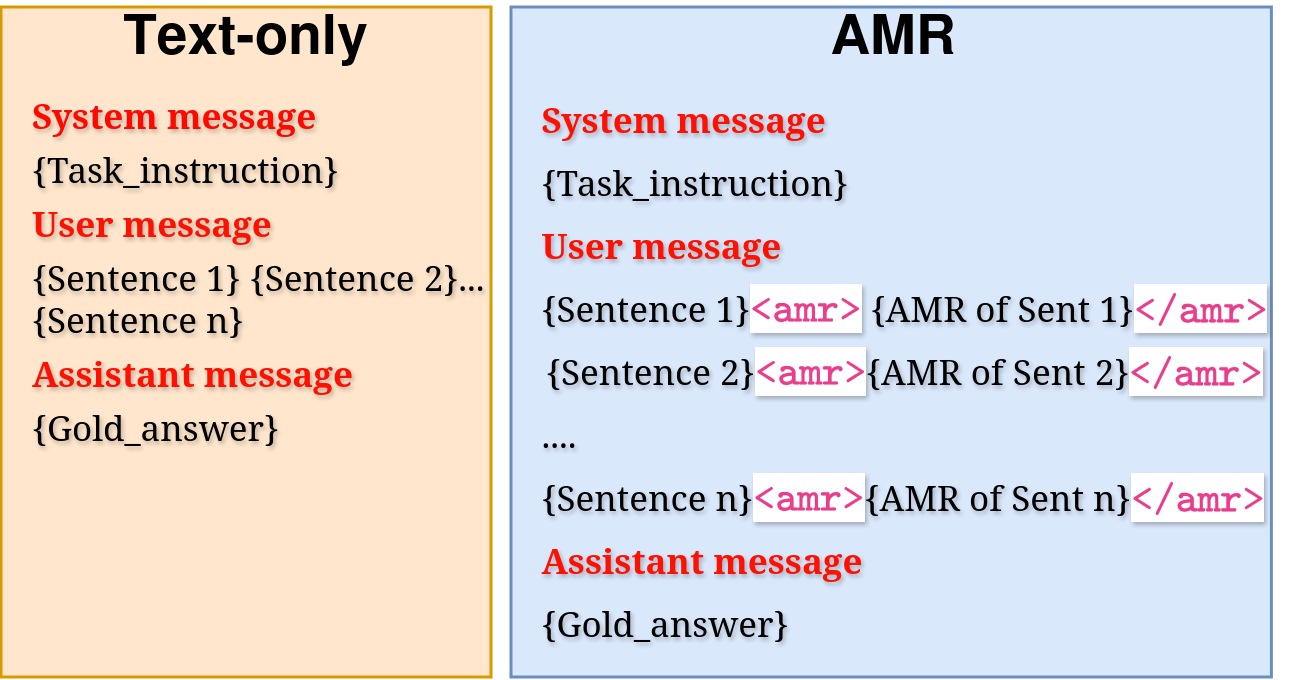} 
  \caption{Templates for text-only (left) and AMR-augmented (right) prompts for the multi-sentence tasks.
 }
  \label{fig:multi-sentence-prompt-template}
\end{figure}

\paragraph{AMR-Text Integration.}
To incorporate AMR structures into the multi-sentence inputs, we interleaved the AMR into the text: each linearized AMR graph is placed immediately after its corresponding sentence, encapsulated within special delimiter tokens (\texttt{<amr>} \dots \texttt{</amr>}) to clearly distinguish AMR from natural language content (see Figure~\ref{fig:multi-sentence-prompt-template}). Full example prompts are provided in Appendix \ref{multi-sentence-appendix}.


\paragraph{Experimental Setup.} The setup follows that of the single-sentence experiments of Section \ref{sec_1sent}. As in Section \ref{sec_1sent}, we experimented with joint and individual-task fine-tuning regimens, and evaluated standard fine-tuning along with a two-phase, intermediate AMR-to-text approach (see Section \ref{sec_1sent_sub_intermediate}).

\subsection{Results}

The results of these experiments are given in Table \ref{tab:multi-sent-downstream-results}. Consistent with our single-sentence findings, we observe no meaningful improvement from AMR augmentation across all fine-tuning strategies and models: the performance of the AMR-augmented models remains on par with or below that of the text-only baseline across all four tasks. 

\begin{table}[t]
\centering
\resizebox{\columnwidth}{!}{%
\begin{tabular}{l l l r r r}
\toprule
\textbf{Dataset} & \textbf{Task} & \textbf{Train} & \textbf{Dev} & \textbf{Test} \\
\midrule
RAMS    & Event Argument Extraction     & 7000 & 849 & 817 \\
CNN/DM  & Abstractive Summarization   & 5000 & 500 & 500 \\
ANLI    & Natural Language Inference     & 7000 & 700 & 1200 \\
LogiQA  & Reading Comprehension        & 3828 & 478 & 480 \\
\bottomrule
\end{tabular}%
}
\caption{Split sizes for the multi-sentence task datasets.}
\label{tab:multisentence-datasets}
\end{table}

Furthermore, note that the performance of the baseline, text-only models on these tasks is not near saturation (perhaps with the exception of ANLI). This indicates that the observed failure in Section \ref{sec_1sent} of AMR augmentation to improve LLMs' downstream performance was not due to saturation on those tasks: across varying model types, fine-tuning regimens, and task difficulties, AMR augmentation consistently fails to markedly improve the performance of modern LLMs.

\begin{table*}[t]
\centering
\small
\begin{tabular}{lllcccc}
\toprule
\textbf{Model} & \textbf{Training} & \textbf{Setup} & \textbf{RAMS} & \textbf{LogiQA} & \textbf{ANLI} & \textbf{CNN} \\
 & & & (F1) & (Acc.) & (F1) & (ROUGE-L) \\
\midrule
\multirow{6}{*}{Llama-3.1-8B}
 & \multirow{3}{*}{Joint}
   & Text          & \textbf{50.0}$_{\pm0.6}$ & \textbf{54.9}$_{\pm0.5}$ & \textbf{88.5}$_{\pm0.5}$ & \textbf{31.5}$_{\pm0.4}$ \\
 & & +AMR          & 49.1$_{\pm0.5}$ & 53.6$_{\pm0.5}$ & 88.1$_{\pm0.4}$ & 31.2$_{\pm0.4}$ \\
 & & +AMR +Inter.  & 49.5$_{\pm0.7}$ & 53.8$_{\pm0.5}$ & 86.8$_{\pm0.6}$ & 31.4$_{\pm0.4}$ \\
\cmidrule{2-7}
 & \multirow{3}{*}{Individual}
   & Text          & 49.4$_{\pm0.7}$ & 55.7$_{\pm1.5}$ & \textbf{87.7}$_{\pm0.9}$ & \textbf{31.0}$_{\pm0.7}$ \\
 & & +AMR          & \textbf{50.1}$_{\pm0.6}$ & \textbf{57.2}$_{\pm0.7}$ & 87.4$_{\pm0.5}$ & 30.7$_{\pm0.5}$ \\
 & & +AMR +Inter.  & 49.9$_{\pm0.4}$ & 53.1$_{\pm1.1}$ & 86.0$_{\pm0.6}$ & 30.7$_{\pm0.4}$ \\
\midrule
\multirow{6}{*}{Qwen3-8B}
 & \multirow{3}{*}{Joint}
   & Text          & 44.2$_{\pm0.8}$ & \textbf{70.7}$_{\pm0.5}$ & \textbf{87.2}$_{\pm0.6}$ & \textbf{30.6}$_{\pm0.5}$ \\
 & & +AMR          & 44.5$_{\pm0.6}$ & 68.8$_{\pm0.7}$ & 86.7$_{\pm0.4}$ & \textbf{30.6}$_{\pm0.5}$ \\
 & & +AMR +Inter.  & \textbf{44.7}$_{\pm0.4}$ & 67.1$_{\pm1.1}$ & 86.3$_{\pm0.5}$ & 30.2$_{\pm0.4}$ \\
\cmidrule{2-7}
 & \multirow{3}{*}{Individual}
   & Text          & \textbf{48.2}$_{\pm0.7}$ & \textbf{70.9}$_{\pm0.7}$ & \textbf{88.0}$_{\pm0.6}$ & \textbf{30.7}$_{\pm0.4}$ \\
 & & +AMR          & \textbf{48.2}$_{\pm0.2}$ & 69.9$_{\pm0.9}$ & 87.8$_{\pm0.5}$ & \textbf{30.7}$_{\pm0.5}$ \\
 & & +AMR +Inter.  & 47.5$_{\pm1.1}$ & 68.6$_{\pm1.1}$ & 86.2$_{\pm0.5}$ & 30.6$_{\pm0.4}$ \\
\bottomrule
\end{tabular}
\caption{Joint and individual training results on the four multi-sentence
task datasets for Llama-3.1-8B-Instruct and Qwen3-8B. Each cell reports
mean $\pm$ standard deviation over five runs with different random seeds.
The best-performing setup within each model, training type, and dataset is
indicated in \textbf{bold}.}
\label{tab:multi-sent-downstream-results}
\end{table*}





\section{Relational Knowledge Analysis}
\label{sec_perplexity}

As the AMR representation $A_S$ of a sentence $S$ is derived solely from $S$ itself\textemdash without the use of any other external source of information\textemdash $A_S$ cannot possibly encode any more information than $S$ already contains. The utility of AMR lies instead in its explicit representation of structure: by making semantic relationships between entities overt and symbolic, AMR has the potential to help a model better extract the latent relational content already present in the text.

The observed failure of AMR augmentation to meaningfully improve downstream performance in the experiments of Sections \ref{sec_1sent}-\ref{sec_multisent} therefore raises the question as to whether modern LLMs require the structural guidance provided by AMR to infer semantic relationships encoded in the text. We posit that AMR does not improve the downstream performance of a modern LLM because it does not provide any relevant relational information not already available to the model.

\begin{table}[t]
\centering
\scalebox{0.67}{
\begin{tabularx}{0.7\textwidth}{l X}
\hline
\textbf{Input Type} & \textbf{User Prompt} \\
\hline
Text-only &
\texttt{Input Sentence: ``The NBA season of 1975--76 was the 30th season of the National Basketball Association.''}\\
\hline
Text+AMR &
\texttt{Input Sentence: ``The NBA season of 1975--76 was the 30th season of the National Basketball Association.''} \newline
\texttt{Input AMR:\newline
(s / season} \newline
\texttt{\hspace*{1em}:ord (o / ordinal-entity} \newline
\texttt{\hspace*{2em}:value 30)} \newline
\texttt{\hspace*{1em}:poss (l / league} \newline
\texttt{\hspace*{2em}:name (n / name} \newline
\texttt{\hspace*{3em}:op1 ...)))} \\
\hline
Text+AMR-nodes &
\texttt{Input Sentence: ``The NBA season of 1975--76 was the 30th season of the National Basketball Association.''} \newline
\texttt{Supplement: season, ordinal-entity, 30, league, name, National, Basketball, Association, date-interval, date-entity, 1975, date-entity, 76}\\
\hline
\end{tabularx}
}
\caption{Input types and their corresponding user prompts. AMR is truncated for presentability. Full example prompts are provided in Appendix~\ref{prompt_for_ppl}.}
\label{tab:ppl-user-prompt}
\end{table}

In this section, we directly evaluate this conjecture (Section \ref{sec_perplexity_sub_method}), and find evidence in support of our hypothesis that AMR augmentation does not improve LLMs' understanding of sentence-level relational knowledge (Section \ref{sec_perplexity_sub_results}).





\subsection{Methodology}
\label{sec_perplexity_sub_method}

Given a sentence $S$ with its corresponding AMR $A_S$, let $R_S$ denote a representation of the relational content of $S$ that is encoded by $A_S$: an encoding of the properties of entities and relations between entities that are expressed by $S$ and represented in $A_S$. If AMR augmentation does in fact provide an LLM $M$ with additional understanding of the relational content of $S$ that $M$ does not already possess, we would expect that the probability assigned to $R_S$ by $M$ would be substantially higher when given both $S$ and its AMR $A_S$, than when given $S$ alone (Equation \ref{eq_amr_prob}).

\begin{equation}
\label{eq_amr_prob}
    P_M(R_S\hspace{1mm}|\hspace{1mm}S,A_S)\gg P_M(R_S\hspace{1mm}|\hspace{1mm}S)
\end{equation}

On the other hand, if our hypothesis holds and LLMs already implicitly encode all of the relational information provided by AMR, we would expect that $P_M(R_S\hspace{1mm}|\hspace{1mm}S,A_S)\approx P_M(R_S\hspace{1mm}|\hspace{1mm}S)$.

\subsubsection{AMR-NLD}
\label{sec_perplexity_sub_method_sub_nld}

As the relational content of an AMR graph is a structured symbolic object, it cannot be directly fed to a language model for likelihood evaluation. To empirically test our hypothesis, it is therefore necessary to transform the linearized AMR graphs of Sections \ref{sec_1sent}-\ref{sec_multisent} into \emph{AMR natural language descriptions} \citep[AMR-NLD;][]{zhang-etal-2025-sr}: natural language descriptions of the relational content of an AMR graph. 

\begin{figure*}[t] 
  \centering
  \includegraphics[width=0.95\textwidth]{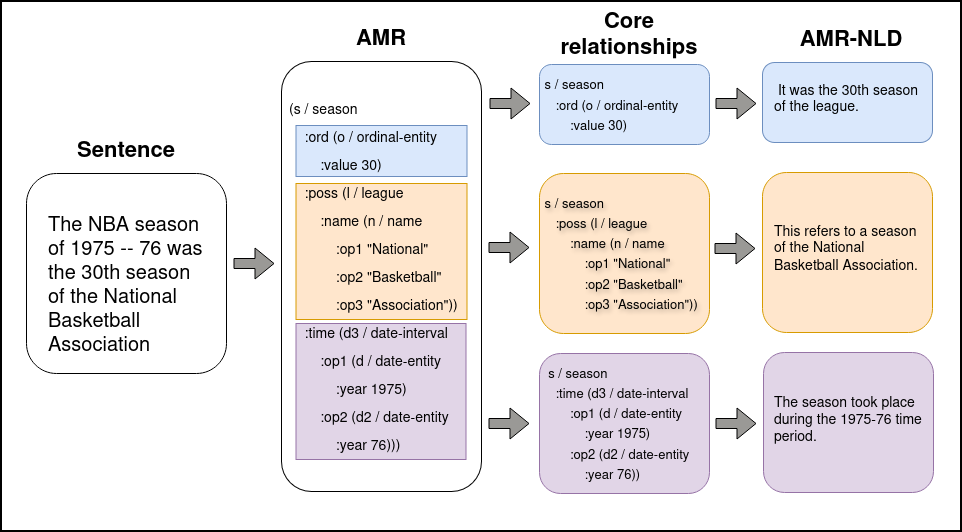} 
  \caption{Our AMR-NLD generation pipeline, illustrated with an example sentence. The AMR graph for the input sentence is decomposed into core relational branches (e.g. :ord, :poss, :time), each of which is mapped to an individual description in natural language.}
  \label{fig:amr-nld}
\end{figure*}

The AMR-NLD generation procedure that we employed is illustrated in Figure \ref{fig:amr-nld}: we first decomposed the AMR into core relational branches, then mapped each to a natural language sentence using Claude Sonnet 4.6 \citep{anthropic2025claude}\textemdash the exact prompt used is provided in Appendix \ref{amr-nld-prompt}. 

\subsubsection{Experimental Setup}
\label{sec_perplexity_sub_method_sub_setup}


We first randomly sampled 300 AMR-sentence pairs and then filtered to 178 pairs (Appendix~\ref{nld-data-construction}) from the PAWS test split used in our experiments in Section \ref{sec_1sent}, and used the procedure described in Section \ref{sec_perplexity_sub_method_sub_nld} to generate the AMR-NLD for each AMR, resulting in a dataset of 178 text, AMR, AMR-NLD triples $(S_i,A_{S_i},R_{S_i})$. 

We then evaluated a series of Llama-3.1-8B and Qwen3-8B models (see below), computing the perplexity of the AMR-NLD $R_{S_i}$ for each example conditioned on (i) the text $S_i$ alone and (ii) $S_i$ and its AMR $A_{S_i}$.

The addition of the AMR $A_{S_i}$ may have a priming effect: it is possible that, for example, the formal, structured AMR representation increases the likelihood that the model assigns to short, declarative sentences such as AMR-NLD\textemdash regardless of the informational content of the AMR. To control for this possibility, we additionally evaluated the models on a third condition, in which we computed the perplexity of the AMR-NLD $R_{S_i}$ conditioned on $S_i$ and a stripped version of the AMR $A_{S_i}$ in which all relations have been removed, leaving only node (entity) labels (\textit{AMR-nodes}; see Table \ref{tab:ppl-user-prompt}).

For each model family (Llama-3.1-8B-Instruct and Qwen3-8B), we evaluated 
four model types: (i) a baseline, off-the-shelf model with no additional fine-tuning; (ii) a model fine-tuned on the text-only version of the PAWS train split (i.e. the original data); (iii) a model fine-tuned on the AMR-augmented PAWS train split following our approach in Section \ref{sec_1sent_sub_method_sub_fting}; and (iv) a model intermediate-fine-tuned on AMR-to-text translation, then fine-tuned on AMR-augmented PAWS, as in our approach outlined in Section \ref{sec_1sent_sub_intermediate}. 

For types (ii)--(iv), the models were each trained with five different random seeds: we report the mean and standard deviation across all five seeds.

\begin{table}[t]
\centering
\small
\begin{tabular}{llll}
\toprule
\textbf{Model} & \textbf{Text} & \textbf{+AMR} & \textbf{+AMR-Nodes} \\
\midrule
\multicolumn{4}{l}{\textit{Llama}} \\
\midrule
Base        & $4.42$ & $3.64$ & $3.72$ \\
Text-FT        & $3.58_{\pm 0.09}$ & $3.06_{\pm 0.11}$ & $3.16_{\pm 0.13}$ \\
AMR-FT         & $3.66_{\pm 0.10}$ & $3.24_{\pm 0.17}$ & $3.33_{\pm 0.10}$ \\
Inter-FT   & $4.66_{\pm 0.15}$ & $4.20_{\pm 0.14}$ & $4.24_{\pm 0.13}$ \\
\midrule
\multicolumn{4}{l}{\textit{Qwen}} \\
\midrule
Base & $19.11$ & $14.91$ & $14.34$ \\
Text-FT       & $5.10_{\pm 0.22}$ & $4.47_{\pm 0.31}$ & $4.55_{\pm 0.25}$ \\
AMR-FT        & $5.07_{\pm 0.30}$ & $4.29_{\pm 0.35}$ & $4.37_{\pm 0.33}$ \\
Inter-FT   & $3.98_{\pm 0.06}$ & $3.79_{\pm 0.04}$ & $3.79_{\pm 0.04}$ \\
\bottomrule
\end{tabular}
\caption{Mean perplexity ($\pm$ std) across model configurations and input conditions.}
\label{tab:perplexity}
\end{table}

\subsection{Results}
\label{sec_perplexity_sub_results}


Perplexity scores by model configuration and input type are given in Table \ref{tab:perplexity}.

AMR augmentation consistently decreases perplexity for all model configurations, seemingly indicating that AMR does in fact provide the LLMs with additional relational information. However, there is no substantial difference between AMR-augmented perplexity and that of the AMR-nodes control condition: recall that this control condition consists of AMR structures that have been stripped of all relational information. This indicates that the observed differences in perplexity between the text-only and AMR-augmented conditions are due to spurious factors\textemdash for example the association of formal-logical structures with short, matter-of-fact statements.

These results therefore support the hypothesis that AMR does not supply relational knowledge beyond what LLMs already infer from text, which is consistent with our results in Sections \ref{sec_1sent} and \ref{sec_multisent}: AMR augmentation offers no benefit to those models on downstream tasks.



\section{Conclusion}

In this paper, we investigated whether augmenting LLM inputs with linearized AMR improves downstream performance. Across two model families, thirteen tasks spanning single- and multi-sentence settings, and several fine-tuning strategies, we found no consistent improvement over text-only baselines: this indicates that the AMR-augmented performance gains reported in prior work are likely due to specific experimental conditions or hyperparameter selection.

We then investigated why AMR augmentation does not benefit modern LLMs. A perplexity-based analysis indicated that AMR does not reduce an LLM's uncertainty about the relational content of a sentence beyond the structure-free control. In other words, AMR augmentation does not provide relational knowledge beyond what is already extractable from text alone for modern LLMs.

\section*{Limitations}

Our multi-seed experiments are limited to 8B parameter models (we conduct a single-seed 70B experiment in Appendix \ref{app:70b}); future work should assess the generalization of our findings to models of different sizes, and in particular examine whether there exists a relationship between model size and the effectiveness of SR integration.

We focus exclusively on the AMR semantic representation format. However, while AMR is the most popular and well-resourced SR format
\citep{wein-opitz-2024-survey}, it is but one of many such formalisms: other SRs such as UCCA and UDS may interact differently with LLMs. Future work should extend our experiments to a broader range of semantic representations.

Additionally, we do not study the potential effect of parser errors on the AMR-augmented models. While the observed performance of AMR-augmented models might be affected by parser errors, we argue that any AMR integration technique will also confront this same issue: it is an inherent limitation of AMR augmentation.

Finally, our study is limited to AMR integration at the fine-tuning stage, and there remains an open question as to whether semantic representations could be more effectively integrated at a more fundamental level such as during pretraining. 


\bibliography{anthology,custom}

\clearpage
\appendix

\section{Single-sentence Experiment} \label{single-sentence-appendix}

\subsection{Dataset construction} \label{single-data-split}

Table \ref{tab:sr-llm-comparison} shows data statistics for 10 datasets used in \citet{zhang-etal-2025-sr}'s work and 9 datasets used in our study. Note that \citet{zhang-etal-2025-sr} do not mention the usage of validation set in their experiment. 

\begin{table*}[t]
\centering
\small
\begin{tabular}{|l|l|r|r|r|r|r|r|}
\hline
\textbf{Dataset} & \textbf{Task} & \multicolumn{3}{c|}{\textbf{Zhang et al. 2025}} & \multicolumn{3}{c|}{\textbf{Our Experiment}} \\
\cline{3-8}
 &  & \textbf{Train} & \textbf{Val} & \textbf{Test} & \textbf{Train} & \textbf{Val} & \textbf{Test} \\
\hline
PAWS & Paraphrase Detection & 10,000 & N/A & 8,000 & 10,000 & 1,000 & 8,000 \\
\hline
SNLI & Textual Entailment & 10,000 & N/A & 10,000 & 10,000 & 1,000 & 10,000 \\
\hline
WMT16 & Translation & 10,000 & N/A & 5,999 & 10,000 & 1,000 & 5,999 \\
\hline
CoNLL2003 & Named Entity Recog. & 10,000 & N/A & 3,453 & 10,000 & 1,000 & 3,453 \\
\hline
LOGIC & Logical Fallacy Det. & 10,000 & N/A & 2,449 & N/A & N/A & N/A \\
\hline
SST-2 & Sentiment Analysis & 10,000 & N/A & 872 & 10,000 & 1,000 & 872 \\
\hline
Pubmed45 & Event Extraction & 10,000 & N/A & 5,000 & 10,000 & 1,000 & 5,000 \\
\hline
WiC & Lexical Disambig. & 5,066 & N/A & 1,048 & 4,418 & 648 & 1,048 \\
\hline
SPIDER & Text2SQL Code Gen. & 7,000 & N/A & 1,034 & 6,300 & 700 & 1,034 \\
\hline
AGNEWS & Text Classification & 10,000 & N/A & 7,600 & 10,000 & 1,000 & 7,600 \\
\hline
\end{tabular}
\caption{Data statistics for \citet{zhang-etal-2025-sr}'s fine-tuning experiment and our fine-tuning experiment}
\label{tab:sr-llm-comparison}
\end{table*}

\paragraph{PAWS, WMT16, Pubmed45, SNLI, CoNLL2003, SST-2, AGNEWS} These datasets contain a large number of records. Following \citet{zhang-etal-2025-sr}, we randomly sample 10,000 examples from each as the training set. Note that since the random seed used in \citet{zhang-etal-2025-sr} was not published, the degree of overlap between the sampled examples in the two studies remains unknown.
\paragraph{LOGIC} \citet{zhang-etal-2025-sr} synthetically generated 10,000 logic examples using GPT-4o-turbo to construct the training set for LOGIC. As the prompts used for data generation were not released, we were unable to reconstruct this dataset and therefore exclude LOGIC from our experiments.
\paragraph{SPIDER} The official SPIDER test set is not publicly available. \citet{zhang-etal-2025-sr} used the original 7,000 training examples for training and the 1,034 validation examples as the test set. We follow the same test set split, but partition the 7,000 training examples into 6,300 for training and 700 for validation.
\paragraph{WiC} \citet{zhang-etal-2025-sr} augmented the original training set with 648 validation examples, yielding a total of 5,066 training samples. We instead retain the original training and validation splits without modification.

\subsection{Examples Prompt }
\label{single_sent_prompts}

In this section, we present complete examples of text-only prompts for all tasks in our single-sentence fine-tuning experiments. AMR-based prompts follow the same structure, with minor modifications to the system message 
to introduce the AMR formalism. AMR prompt examples are available in our GitHub repository.

\paragraph{agnews}
\textit{[system]} You are an expert in news article classification for the AG News dataset. Your goal is to categorize news articles into one of four topics based on their content. You will be given a news article text. Analyze the content carefully and determine which category it belongs to. Respond only with one of these four labels: World, Sports, Business, or Sci/Tech. Do not include explanations, punctuation, or any additional text in your output.

\textit{[user]} Unions representing workers at Turner Newall say they are `disappointed' after talks with stricken parent firm Federal Mogul.

\textit{[assistant]} Business

\paragraph{conll}
\textit{[system]} You are an expert in Named Entity Recognition for the CoNLL-2003 dataset. Your goal is to identify and classify named entities in text. You will be given a sentence and a list of its tokens. For each token, assign the appropriate NER tag using the IOB2 format. The tag set includes: B-PER (beginning of person name), I-PER (inside person name), B-ORG (beginning of organization), I-ORG (inside organization), B-LOC (beginning of location), I-LOC (inside location), B-MISC (beginning of miscellaneous entity), I-MISC (inside miscellaneous entity), and O (outside any named entity). Respond only with a list of tags in the same order as the input tokens, formatted as a Python list (e.g., \texttt{['B-LOC', 'O', 'B-PER']}). Do not include explanations, punctuation, or any additional text in your output.

\textit{[user]} Japan began the defence of their Asian Cup title with a lucky 2-1 win against Syria in a Group C championship match on Friday .

\textit{[assistant]} \texttt{['B-LOC', 'O', 'O', 'O', 'O', 'O', 'B-MISC', 'I-MISC', 'O', 'O', 'O', 'O', 'O', 'O', 'O', 'B-LOC', 'O', 'O', 'O', 'O', 'O', 'O', 'O', 'O', 'O']}

\paragraph{paws}
\textit{[system]} You are an expert in paraphrase identification for the PAWS dataset. Your goal is to determine whether two sentences express the same meaning despite differences in wording or structure. You will be given two sentences. Analyze their semantics carefully and determine whether the two sentences convey the same meaning. Respond only with one of these two labels: Yes or No. Do not include explanations, punctuation, or any additional text in your output.

\textit{[user]} Sentence 1: It is situated south of K\"{o}ro\u{g}lu Mountains and to the north of Bolu. Sentence 2: It is situated south of K\"{o}ro\u{g}lu-mountains and north of the Bolu.

\textit{[assistant]} Yes

\paragraph{pubmed}
\textit{[system]} You are an expert in biomedical relation extraction for the PubMed 45 dataset. Your goal is to validate protein interaction tuples extracted from biomedical literature. You will be given an interaction tuple and the source sentence from which it was extracted. The interaction tuple contains an interaction type, a catalyst protein, and one or two other proteins involved in the interaction. Analyze whether the tuple correctly represents a valid interaction as described in the sentence. Respond only with one of these three labels: 0, 1, or 2. Label 1 means the interaction tuple is valid. Label 0 means it is invalid. Label 2 means that swapping the catalyst role with one of the other proteins would make the interaction valid. Do not include explanations, punctuation, or any additional text in your output.

\textit{[user]} Sentence: High molecular mass species containing CCR5 and CXCR4 simultaneously are not detected, as shown when cells are stimulated with SDF-1$\alpha$ or RANTES and (AOP)-RANTES, and when immunoprecipitation and Western blot are performed using anti-CCR5 or anti-CXCR4 antibodies, respectively. Interaction tuple: \texttt{['stimulate', 'AOP', 'CXCR4']}

\textit{[assistant]} 0

\paragraph{snli}
\textit{[system]} You are an expert in the Natural Language Inference task for the SNLI dataset. Your goal is to determine the logical relationship between a premise and a hypothesis. You will be provided with a premise and a hypothesis statement. Carefully analyze their meanings and classify their relationship into one of the following three categories: entailment (the hypothesis is definitely true given the premise), neutral (the hypothesis might be true but is not guaranteed by the premise), or contradiction (the hypothesis is false or incompatible with the premise). Respond only with one of these three labels: entailment, neutral, or contradiction. Do not include explanations, punctuation, or any additional text in your output.

\textit{[user]} Premise: This church choir sings to the masses as they sing joyous songs from the book at a church. Hypothesis: The church has cracks in the ceiling.

\textit{[assistant]} neutral

\paragraph{spider}
\textit{[system]} You are an expert in text-to-SQL generation for the Spider dataset. Your goal is to convert natural language questions into accurate SQL queries. You will be given a natural language question and a database schema. Write an SQL query that retrieves the requested information based on the given question. Use proper SQL syntax and consider any necessary table joins, conditions, aggregations, and sorting operations. Respond only with the SQL query.

\textit{[user]} Question: How many singers do we have? Database schema: \{database\_schema\}

\textit{[assistant]} \texttt{SELECT count(*) FROM singer}

\paragraph{sst-2}
\textit{[system]} You are an expert in sentiment analysis for the SST-2 dataset. Your goal is to determine the sentiment expressed in movie reviews. You will be given a sentence from a movie review. Analyze the sentiment carefully and determine whether it expresses a positive or negative opinion. Respond only with one of these two labels: positive or negative. Do not include explanations, punctuation, or any additional text in your output.

\textit{[user]} it 's a charming and often affecting journey .

\textit{[assistant]} positive

\paragraph{wic}
\textit{[system]} You are an expert in word sense disambiguation for the WiC dataset. Your goal is to identify whether a target word has the same meaning across two different contexts. You will be given a target word (either a noun or a verb) and two sentences, each containing that word. Analyze whether the target word is used with the same meaning in both contexts. Respond only with one of these two labels: True or False. True means the word has the same meaning in both contexts. False means the word has different meanings in the two contexts. Do not include explanations, punctuation, or any additional text in your output.

\textit{[user]} Sentence 1: It was a narrow defeat. Sentence 2: The army's only defeat. Target: defeat

\textit{[assistant]} True

\paragraph{wmt}
\textit{[system]} You are an expert in machine translation for the WMT dataset. Your goal is to translate English sentences into German accurately and fluently. You will be given an English sentence. Translate the sentence into German, ensuring that the translation preserves the meaning, style, and nuances of the original text. Respond only with the German translation.

\textit{[user]} Obama receives Netanyahu

\textit{[assistant]} Obama empf\"{a}ngt Netanyahu

\subsection{Fine-tuning Configuration}
We perform a grid search to select the optimal learning rate for each  configuration, using validation loss as the selection criterion. 
For joint fine-tuning, we search over five candidate values: 
$\{1e{-5}, 3e{-5}, 5e{-5}, 7e{-5}, 1e{-4}\}$. 
For individual fine-tuning, due to the large number of runs, we reduce the search pool to three values: 
$\{1e{-5}, 5e{-5}, 1e{-4}\}$. 
Tables~\ref{tab:lr_llama} and~\ref{tab:lr_qwen} report the best 
learning rate found for LLaMa and Qwen, respectively. 

The rest of training configuration are presented in Table~\ref{tab:single-sentence-rest-details}.

\begin{table}[h]
\centering
\small
\begin{tabular}{llcc}
\toprule
\textbf{Train} & \textbf{Dataset} & \textbf{Setup} & \textbf{LR} \\
\midrule
\multirow{27}{*}{Indiv.}
 & \multirow{3}{*}{PubMed}  & Text    & $5e{-5}$ \\
 &                          & AMR  & $5e{-5}$ \\
 &                          & AMR-i.  & $5e{-5}$ \\
 & \multirow{3}{*}{WMT}     & Text    & $5e{-5}$ \\
 &                          & AMR  & $5e{-5}$ \\
 &                          & AMR-i.  & $5e{-5}$ \\
 & \multirow{3}{*}{PAWS}    & Text    & $5e{-5}$ \\
 &                          & AMR  & $5e{-5}$ \\
 &                          & AMR-i.  & $5e{-5}$ \\
 & \multirow{3}{*}{WiC}     & Text    & $5e{-5}$ \\
 &                          & AMR  & $5e{-5}$ \\
 &                          & AMR-i.  & $5e{-5}$ \\
 & \multirow{3}{*}{CoNLL}   & Text    & $5e{-5}$ \\
 &                          & AMR  & $5e{-5}$ \\
 &                          & AMR-i.  & $5e{-5}$ \\
 & \multirow{3}{*}{SST}     & Text    & $5e{-5}$ \\
 &                          & AMR  & $5e{-5}$ \\
 &                          & AMR-i.  & $5e{-5}$ \\
 & \multirow{3}{*}{Spider}  & Text    & $5e{-5}$ \\
 &                          & AMR  & $5e{-5}$ \\
 &                          & AMR-i.  & $5e{-5}$ \\
 & \multirow{3}{*}{AGNews}  & Text    & $5e{-5}$ \\
 &                          & AMR  & $5e{-5}$ \\
 &                          & AMR-i.  & $5e{-5}$ \\
 & \multirow{3}{*}{SNLI}    & Text    & $5e{-5}$ \\
 &                          & AMR  & $5e{-5}$ \\
 &                          & AMR-i.  & $5e{-5}$ \\
\midrule
\multirow{3}{*}{Joint}
 & -- & Text   & $7e{-5}$ \\
 & -- & AMR. & $5e{-5}$ \\
 & -- & AMR-i. & $5e{-5}$ \\
\bottomrule
\end{tabular}
\caption{Learning rates for Llama per training configuration. AMR-i. stands for AMR-Intermediate. Dataset names are simplified for illustration purposes.}
\label{tab:lr_llama}
\end{table}

\begin{table}[h]
\centering
\small
\begin{tabular}{llcc}
\toprule
\textbf{Train} & \textbf{Dataset} & \textbf{Setup} & \textbf{LR} \\
\midrule
\multirow{27}{*}{Indiv.}
 & \multirow{3}{*}{PubMed}  & Text    & $5e{-5}$ \\
 &                          & AMR  & $5e{-5}$ \\
 &                          & AMR-i.  & $5e{-5}$ \\
 & \multirow{3}{*}{WMT}     & Text    & $5e{-5}$ \\
 &                          & AMR  & $5e{-5}$ \\
 &                          & AMR-i.  & $5e{-5}$ \\
 & \multirow{3}{*}{PAWS}    & Text    & $5e{-5}$ \\
 &                          & AMR  & $5e{-5}$ \\
 &                          & AMR-i.  & $5e{-5}$ \\
 & \multirow{3}{*}{WiC}     & Text    & $5e{-5}$ \\
 &                          & AMR  & $5e{-5}$ \\
 &                          & AMR-i.  & $5e{-5}$ \\
 & \multirow{3}{*}{CoNLL}   & Text    & $5e{-5}$ \\
 &                          & AMR  & $5e{-5}$ \\
 &                          & AMR-i.  & $5e{-5}$ \\
 & \multirow{3}{*}{SST}     & Text    & $5e{-5}$ \\
 &                          & AMR  & $5e{-5}$ \\
 &                          & AMR-i.  & $5e{-5}$ \\
 & \multirow{3}{*}{Spider}  & Text    & $5e{-5}$ \\
 &                          & AMR  & $5e{-5}$ \\
 &                          & AMR-i.  & $5e{-5}$ \\
 & \multirow{3}{*}{AGNews}  & Text    & $5e{-5}$ \\
 &                          & AMR  & $5e{-5}$ \\
 &                          & AMR-i.  & $5e{-5}$ \\
 & \multirow{3}{*}{SNLI}    & Text    & $5e{-5}$ \\
 &                          & AMR  & $5e{-5}$ \\
 &                          & AMR-i.  & $5e{-5}$ \\
\midrule
\multirow{3}{*}{Joint}
 & -- & Text   & $7e{-5}$ \\
 & -- & AMR. & $5e{-5}$ \\
 & -- & AMR-i. & $5e{-5}$ \\
\bottomrule
\end{tabular}
\caption{Learning rates for Qwen per training configuration. AMR-i. stands for AMR-Intermediate.}
\label{tab:lr_qwen}
\end{table}

\begin{table}[h]
\centering
\small
\scalebox{0.99}{\begin{tabular}{@{}p{2.8cm}p{4.5cm}@{}}
\toprule
\textbf{Parameter} & \textbf{Value} \\
\midrule
\multicolumn{2}{l}{\textit{LLM Related}} \\
Attention impl. & \texttt{flash\_attention\_2} \\
Dtype & \texttt{bfloat16} \\
\midrule
\multicolumn{2}{l}{\textit{LoRA Configuration}} \\
Rank ($r$)    & 64 \\
Alpha         & 128 \\
Dropout       & 0.05 \\
Target modules & \texttt{q\_proj, k\_proj, v\_proj, o\_proj, gate\_proj, up\_proj, down\_proj} \\
\midrule
\multicolumn{2}{l}{\textit{Training}} \\
Random seed         & 0, 1, 2, 3, 4 \\
Batch size          & 8 \\
Weight decay        & 0.01 \\
Early stopping patience & 3 \\
Grad. accum. steps  & 4 \\
Training epochs     & 10 \\
\midrule
\multicolumn{2}{l}{\textit{Inference}} \\
Max new tokens & 50 \\
Do sample      & False \\
\multicolumn{1}{l}{\textit{GPU}} \\
Number of GPUs & 1\\
\bottomrule
\end{tabular}}
\caption{Configuration for single-sentence fine-tuning.}
\label{tab:single-sentence-rest-details}
\end{table}

\subsection{Scaling to 70B}
\label{app:70b}

Due to compute constraints, we ran each configuration with a single seed and did not evaluate the intermediate
AMR-to-text condition. Results are given in Table~\ref{tab:70b}.

\begin{table*}[t]
\centering
\small
\begin{tabular}{llccccccccc}
\toprule
\textbf{Training} & \textbf{Setup} & \textbf{WiC} & \textbf{SST} & \textbf{SNLI} & \textbf{PM} & \textbf{PAWS} & \textbf{AGN} & \textbf{SPD} & \textbf{CNL} & \textbf{WMT} \\
 & & (F1) & (F1) & (F1) & (F1) & (F1) & (F1) & (EM) & (F1) & (BLEU) \\
\midrule
\multirow{2}{*}{Joint}
 & Text  & 76.2 & 96.1 & \textbf{92.1} & 67.4 & \textbf{93.9} & \textbf{92.6} & 59.1 & \textbf{93.0} & 30.2 \\
 & +AMR  & \textbf{78.1} & \textbf{96.7} & 92.0 & \textbf{73.9} & 93.5 & 92.1 & \textbf{61.5} & 92.4 & \textbf{31.3} \\
\midrule
\multirow{2}{*}{Individual}
 & Text  & 78.0 & 96.7 & \textbf{92.2} & 83.9 & \textbf{94.2} & \textbf{93.6} & \textbf{59.4} & 92.9 & \textbf{31.3} \\
 & +AMR  & \textbf{78.5} & \textbf{97.0} & 92.0 & \textbf{85.2} & 93.7 & 93.2 & 58.8 & \textbf{93.6} & 28.2 \\
\bottomrule
\end{tabular}
\caption{Joint and individual training results on the nine
single-sentence task datasets for Llama-3.1-70B-Instruct.
The best-performing setup within each training type and dataset is
indicated in \textbf{bold}.}
\label{tab:70b}
\end{table*}

\section{AMR-to-text generation} \label{amr2text}

\subsection{Experimental details} \label{amr2text_exp}

Figure~\ref{fig:amr2text_prompt_template} presents the prompt template used for training. The model is loaded in bfloat16 precision and fine-tuned using LoRA with rank $r=64$, $\alpha=128$, and a dropout of 0.05. LoRA adapters are applied to all seven projection layers: query, key, value, output, gate, up, and down projections.

Training runs for 10 epochs with a batch size of 8 on a single GPU. The AdamW optimiser is used with a learning rate of $1\times10^{-4}$ and weight decay of 0.01. Random seed is set to 42. Best model selected based on lowest validation loss.

\begin{figure}[t]
\centering
\small
\begin{tabular}{|p{0.93\columnwidth}|}
\hline
\textbf{\textit{\{System\}}} Generate a sentence for the given Abstract Meaning Representation. \\
\textbf{\textit{\{User\}}} \texttt{(r / receive-01 :ARG0 (p / person :name :op1 ``Obama'') :ARG1 (p2 / person :name ``Netanyahu''))} \\
\textbf{\textit{\{Assistant\}}} Obama receives Netanyahu \\
\hline
\end{tabular}
\caption{Prompt template used for AMR-to-text generation.}
\label{fig:amr2text_prompt_template}
\end{figure}

\subsection{Results} \label{amr2text_result}

We report BLEU scores on the AMR 3.0 test split after intermediate fine-tuning in Table \ref{tab:amr_translation_results}. Our approach enables 8B models to achieve strong AMR-to-text performance after Phase 1, competitive with state-of-the-art model \citep{cheng-etal-2022-bibl}. After Phase 2, performance drops moderately — from 46 to 43.9 BLEU — an expected consequence of catastrophic forgetting when adapting to new tasks. Nevertheless, the retained score of 43.9 BLEU indicates that the model preserves substantial AMR comprehension after Phase 2 training.

\begin{table}[h]
\centering
\small
\begin{tabular}{|l|c|}
\hline
\textbf{Model Strategy} & \textbf{BLEU Score} \\
\hline
\multicolumn{2}{|l|}{\textbf{LLaMA3.1-8B-Ins}} \\
\hline
Phase 1 checkpoint & 46.0 \\
\hline
Phase 2 checkpoint & 43.9 \\
\hline
\multicolumn{2}{|l|}{\textbf{Qwen3-8B}} \\
\hline
Phase 1 checkpoint & 45.7 \\
\hline
Phase 2 checkpoint & 43.1 \\
\hline
\multicolumn{2}{|l|}{\textbf{State-of-the-Art (Reference)}} \\
\hline
~~ BiBL & 47.4 \\
\hline
\end{tabular}
\caption{AMR-to-text generation performance (BLEU) for LLaMA and Qwen models}
\label{tab:amr_translation_results}
\end{table}

\section{Multi-sentence Experiment} \label{multi-sentence-appendix}

\subsection{Dataset Construction}

\paragraph{RAMS}
The original RAMS dataset contains 7329, 924, and 827 samples for the train, development, and test splits, respectively. Each example consists of a passage of 3--5 sentences. We use the complete dataset without additional filtering, except for excluding samples that fail AMR parsing, resulting in 7000, 849, and 817 samples for the train, development, and test splits.

\paragraph{CNN/DailyMail}
The original dataset contains 287k, 13.4k, and 11.5k samples for the train, development, and test splits, with considerable variation in passage length. We first filter out examples with more than 10 sentences, then randomly sample 5000, 500, and 500 samples for the train, development, and test sets, respectively.

\paragraph{ANLI}
The original dataset contains 100k, 1.2k, and 1.2k samples for the train, development, and test splits, respectively. We randomly sample 7000 and 700 samples for the train and development sets.

\paragraph{LogiQA}
The original dataset comprises 7,380, 651, and 651 samples for the train, development, and test splits, respectively. We exclude single-sentence samples as they do not satisfy the multi-sentence requirement of our experimental setup and samples that fail AMR parsing, leaving 3828, 478, and 480 samples for the train, development, and test splits, respectively.

\subsection{Example prompts}
In this section, we present complete examples of text-only prompts for all tasks in our multi-sentence fine-tuning experiments. AMR-based prompts follow the same structure, with minor modifications to the system message 
to introduce the AMR formalism. AMR prompt examples are available in our GitHub repository.

\paragraph{RAMS}
\textit{[system]} You are an expert in event argument extraction task for RAMS dataset. Your task is to find argument for specific events.

\textbf{\#\# Task Definition}

Given a text passage, an event trigger word/phrase, an event type, and all the argument roles associated with that event type, you must extract the exact text spans from the passage that fill each role.

\textbf{\#\# Instructions}
\begin{itemize}[noitemsep, topsep=2pt]
    \item Process each argument role individually.
    \item For each role:
    \begin{itemize}[noitemsep, topsep=2pt]
        \item Find corresponding text span in the passage.
        \item A span must be a contiguous, exact quote from the passage and within a sentence. A span cannot cross sentence boundaries.
        \item If the role cannot be filled based on the given passage, don't include it in your answer.
    \end{itemize}
\end{itemize}

\textbf{\#\# Output Format}

Provide your response as a list of role-span pairs, one per line:

\texttt{role\_name1:exact\_text\_span1}

\texttt{role\_name2:exact\_text\_span2}

\textbf{\#\# Important Notes}
\begin{itemize}[noitemsep, topsep=2pt]
    \item Event types are provided in hierarchical format using dot notation, where each level represents increasing specificity: \texttt{main\_type.sub\_type.sub\_sub\_type}
    \item Do not include explanations or additional commentary.
\end{itemize}

\textit{[user]}
\begin{itemize}[noitemsep, topsep=2pt]
    \item Passage: Three specific points illustrate why Americans see Trump as the problem: 1) Trump has trouble working with people beyond his base. In Saddam Hussein's Iraq that might work when opponents can be thrown in jail or exterminated. In the United States that won't fly: presidents must build bridges within and beyond their core support to resolve challenges. Without alliances, a president can't get approval to get things done.
    \item Event trigger: exterminated
    \item Event type: \texttt{life.die.n/a}
    \item Argument roles: victim, place
\end{itemize}

\textit{[assistant]} \texttt{victim:opponents} \texttt{place:Iraq}

\paragraph{CNN}
\textit{[system]} You are an expert in summarization task for CNN dataset.

\textbf{\#\# Task Definition}

Given a passage, generate a concise and accurate summary. Your response should consist only of the summary --- do not include any introductory or explanatory text.

\textit{[user]}
\begin{itemize}[noitemsep, topsep=2pt]
    \item Passage: Baghdad (CNN) -- Radical Iraqi cleric Muqtada al-Sadr has returned to the country after more than three years in Iran, according to Iraqi state television and websites maintained by al-Sadr's followers. The Shiite cleric has been in Iran since early 2007, apart from a public appearance in Iraq in May 2007. He has been studying in the Iranian city of Qom to become an ayatollah, the title given to high-ranking Shiite Muslim religious scholars. Al-Sadr commanded one of Iraq's most formidable private armies after the fall of Saddam Hussein, which contributed to violence and instability in the country for several years. His political bloc has now joined forces with a former rival, Prime Minister Nuri al-Maliki. The Sadr movement emerged as one of the kingmakers in Iraqi politics in March, when it won 39 parliamentary seats. The bloc's support played a major role in al-Maliki getting his second term in office.
\end{itemize}

\textit{[assistant]} Muqtada al-Sadr has been in Iran since 2007. He's been studying to be an ayatollah. His political bloc was a kingmaker after elections in March.

\paragraph{LogiQA}
\textit{[system]} You are an expert in logical reasoning for LogiQA dataset. You will be provided with a context and a question and multiple answer options. Your task is to select the correct answer. Output only the index number (starting from 0) of the correct option. Do not include explanations, punctuation, or additional text in your output.

\textit{[user]}
\begin{itemize}[noitemsep, topsep=2pt]
    \item Passage: Continuous exposure to indoor fluorescent lights is beneficial to the health of hamsters with heart disease. One group of hamsters exposed to continuous exposure to fluorescent lights has an average lifespan that is 2.5\% longer than another one of the same species but living in a black wall.
    \item Question: Which of the following questions was the initial motivation for conducting the above experiment?
    \item Options: [``Can hospital light therapy be proved to promote patient recovery?'', ``Which one lives longer, the hamster living under the light or the hamster living in the dark?'', ``What kind of illness does the hamster have?'', ``Do some hamsters need a period of darkness?'']
\end{itemize}

\textit{[assistant]} 0

\paragraph{ANLI}
\textit{[system]} You are an expert in the Adversarial Natural Language Inference (ANLI) task. Your goal is to determine the logical relationship between a premise and a hypothesis. You will be provided with a passage as the premise and a hypothesis statement. Carefully analyze their meanings and classify their relationship into one of the following three categories: entailment (the hypothesis is definitely true given the premise), neutral (the hypothesis might be true but is not guaranteed by the premise), or contradiction (the hypothesis is false or incompatible with the premise). Respond only with one of these three labels: entailment, neutral, or contradiction. Do not include explanations, punctuation, or any additional text in your output.

\textit{[user]}
\begin{itemize}[noitemsep, topsep=2pt]
    \item Premise: McGee (lower body) has been cleared for Friday's tilt with the Grizzlies. McGee will play through a glute contusion Friday night, but the injury will definitely be something to keep an eye on. He may see his minutes monitored more closely, in which case expect Damian Jones to see increased action.
    \item Hypothesis: The glute is the injured part of him.
\end{itemize}

\textit{[assistant]} entailment

\subsection{Fine-tuning configuration}
We perform a grid search to select the optimal learning rate for each  configuration, using validation loss as the selection criterion. We search over five candidate values: $\{1e{-5}, 3e{-5}, 5e{-5}, 7e{-5}, 1e{-4}\}$. Tables~\ref{tab:lr_multi} report the best 
learning rate found for Llama and Qwen. 

\begin{table}[t]
\centering
\small
\begin{tabular}{lllcc}
\toprule
\textbf{Train} & \textbf{Model} & \textbf{Dataset} & \textbf{Setup} & \textbf{LR} \\
\midrule
\multirow{24}{*}{Indiv.}
 & \multirow{12}{*}{Llama}
   & \multirow{3}{*}{RAMS}   & Text   & $5e{-5}$\\
 & &                         & AMR & $1e{-5}$\\
 & &                         & AMR-i. & $3e{-5}$\\
 & & \multirow{3}{*}{LogiQA}    & Text  &  $1e{-5}$\\
 & &                         & AMR &  $1e{-5}$\\
 & &                         & AMR-i. &  $3e{-5}$\\
 & & \multirow{3}{*}{ANLI} & Text   & $1e{-5}$\\
 & &                         & AMR & $1e{-5}$\\
 & &                         & AMR-i. & $1e{-5}$\\
 & & \multirow{3}{*}{CNN}   & Text   & $3e{-5}$\\
 & &                         & AMR & $1e{-5}$\\
 & &                         & AMR-i. & $1e{-5}$\\
\cmidrule{2-5}
 & \multirow{12}{*}{Qwen}
   & \multirow{3}{*}{RAMS}   & Text   & $3e{-5}$\\
 & &                         & AMR & $1e{-5}$\\
 & &                         & AMR-i. & $3e{-5}$\\
 & & \multirow{3}{*}{LogiQA}    & Text   &  $1e{-5}$\\
 & &                         & AMR &  $1e{-5}$\\
 & &                         & AMR-i. &  $1e{-5}$\\
 & & \multirow{3}{*}{ANLI} & Text   & $1e{-5}$\\
 & &                         & AMR & $1e{-5}$\\
 & &                         & AMR-i. & $1e{-5}$\\
 & & \multirow{3}{*}{CNN}   & Text   & $1e{-5}$\\
 & &                         & AMR & $3e{-5}$\\
 & &                         & AMR-i. & $1e{-5}$\\
\midrule
\multirow{6}{*}{Joint}
 & \multirow{3}{*}{Llama}
   & -- & Text   & $1e{-5}$\\
 & & -- & AMR & $1e{-5}$\\
 & & -- & AMR-i. & $1e{-5}$\\
\cmidrule{2-5}
 & \multirow{3}{*}{Qwen}
   & -- & Text   & $1e{-5}$\\
 & & -- & AMR & $1e{-5}$\\
 & & -- & AMR-i. & $1e{-5}$\\
\bottomrule
\end{tabular}
\caption{Learning rates per training configuration for RAMS, CNN, LogiQA, and ANLI. AMR-i. stands for AMR-Intermediate}
\label{tab:lr_multi}
\end{table}

The rest of training configuration are presented in Table~\ref{tab:multi-sentence-rest-details}.

\begin{table}[h]
\centering
\small
\scalebox{0.99}{\begin{tabular}{@{}p{2.8cm}p{4.5cm}@{}}
\toprule
\textbf{Parameter} & \textbf{Value} \\
\midrule
\multicolumn{2}{l}{\textit{LLM Related}} \\
Attention impl. & \texttt{flash\_attention\_2} \\
Dtype & \texttt{bfloat16} \\
\midrule
\multicolumn{2}{l}{\textit{LoRA Configuration}} \\
Rank ($r$)    & 64 \\
Alpha         & 128 \\
Dropout       & 0.05 \\
Target modules & \texttt{q\_proj, k\_proj, v\_proj, o\_proj, gate\_proj, up\_proj, down\_proj} \\
\midrule
\multicolumn{2}{l}{\textit{Training}} \\
Random seed         & 0,1,2,3,4 \\
Batch size          & 4 \\
Weight decay        & 0.01 \\
Early stopping patience & 3 \\
Grad. accum. steps  & 4 \\
Training epochs     & 10 \\
\midrule
\multicolumn{2}{l}{\textit{Inference}} \\
Max new tokens & 50 \\
Do sample      & False \\
\multicolumn{1}{l}{\textit{GPU}} \\
Number of GPUs & 1\\
\bottomrule
\end{tabular}}
\caption{Configuration for multi-sentence fine-tuning.}
\label{tab:multi-sentence-rest-details}
\end{table}




\section{Alternative Integration Strategies}
\label{app:alt_integration}

\paragraph{GNN encoder.}
We encode each sentence-level AMR graph with a 3-layer relational graph
convolutional network \citep{schlichtkrull2018modeling} of hidden dimension 768,
using 17 relation types obtained by bucketing the AMR edge labels. Node features are initialized from the
LLM's own input embedding table by mean-pooling the subword embeddings of each
concept label, so the graph encoder and the language model share an input space. To project the resulting node representations into the LLM's embedding space we
use a querying transformer (Q-Former) in the style of BLIP-2
\citep{li2023blip}: $K$ learnable query vectors self-attend and then cross-attend to the node representations of a single graph, yielding a fixed
number of $K$ outputs per graph regardless of graph size, which a final linear
layer maps to the LLM embedding dimension. We place graph tokens immediately after its corresponding sentence. The
graph encoder and Q-Former are trained jointly with the LoRA adapters, while the
base LLM remains frozen.

\paragraph{AMRBART encoder.}
We use the encoder
of AMRBART \citep{bai-etal-2022-graph}, a BART-based sequence-to-sequence model
pre-trained on AMR, to encode the linearized graph. Graphs are serialized in
AMRBART's own format and tokenized with its graph-aware vocabulary. These vectors are then projected into the LLM's embedding space by
exactly the same Q-Former architecture as in GNN encoder approach. The AMRBART encoder is kept frozen and only the Q-Former is trained, alongside the LoRA adapters.

Due to compute constraints, we ran each configuration with a single seed. Results are given in Tables~\ref{tab:alt_single} and
\ref{tab:alt_multi}. Both strategies show no consistent improvement over the text-only baselines reported in Section \ref{sec_1sent} and Section \ref{sec_multisent}. 

\begin{table}[t]
\centering
\small
\setlength{\tabcolsep}{3.5pt}
\begin{tabular}{@{}llcccc@{}}
\toprule
\textbf{Strategy} & \textbf{Training} & \textbf{RAMS} & \textbf{LogiQA} & \textbf{ANLI} & \textbf{CNN} \\
 & & (F1) & (Acc.) & (F1) & (R-L) \\
\midrule
\multirow{2}{*}{GNN}
 & Joint  & 50.2 & 54.6 & 88.8 & 31.0 \\
 & Indiv. & 50.7 & 57.5 & 88.1 & 31.2 \\
\midrule
\multirow{2}{*}{AMRBART}
 & Joint  & 49.5 & 56.9 & 88.7 & 32.1 \\
 & Indiv. & 49.4 & 56.9 & 87.7 & 31.2 \\
\bottomrule
\end{tabular}
\caption{Multi-sentence task results for Llama-3.1-8B-Instruct with
GNN- and AMRBART-encoded AMR soft prompts.}
\label{tab:alt_multi}
\end{table}

\begin{table*}[t]
\centering
\small
\begin{tabular}{llccccccccc}
\toprule
\textbf{Strategy} & \textbf{Training} & \textbf{WiC} & \textbf{SST} & \textbf{SNLI} & \textbf{PM} & \textbf{PAWS} & \textbf{AGN} & \textbf{SPD} & \textbf{CNL} & \textbf{WMT} \\
 & & (F1) & (F1) & (F1) & (F1) & (F1) & (F1) & (EM) & (F1) & (BLEU) \\
\midrule
\multirow{2}{*}{GNN}
 & Joint      & 74.0 & 95.8 & 90.9 & 71.3 & 93.2 & 88.8 & 53.4 & 90.9 & 27.7 \\
 & Individual & 74.7 & 96.2 & 91.0 & 74.4 & 93.5 & 93.0 & 52.1 & 92.4 & 26.6 \\
\midrule
\multirow{2}{*}{AMRBART}
 & Joint      & 72.3 & 96.0 & 91.0 & 81.7 & 93.7 & 74.9 & 52.8 & 93.0 & 26.1 \\
 & Individual & 75.0 & 96.3 & 91.5 & 84.4 & 93.9 & 91.4 & 49.8 & 92.2 & 28.4 \\
\bottomrule
\end{tabular}
\caption{Single-sentence task results for Llama-3.1-8B-Instruct with
GNN- and AMRBART-encoded AMR soft prompts.}\label{tab:alt_single}
\end{table*}

\section{Relational Knowledge Analysis} 
\subsection{Dataset Construction} \label{nld-data-construction}

We first randomly sample 300 sentences from the PAWS dataset. We then filter out sentences whose semantic content is too simple to yield a meaningful AMR-NLD --- for example, sentences such as ``I saw a dog'', whose AMR-NLD would be largely identical to the sentence itself and thus uninformative for our analysis. Filtering is performed using Claude Sonnet 4.6, resulting in a final set of 178 sentences for AMR-NLD generation.

\subsection{Prompt For Relational Knowledge Analysis} \label{prompt_for_ppl}
Table~\ref{tab:ppl-prompt} presents the system prompts used for getting perplexity. 

\begin{table*}[t] 
\centering
{\scriptsize  
\begin{tabularx}{\textwidth}{l >{\hsize=1\hsize}X >{\hsize=1\hsize}X}
\hline
\textbf{Type of Prompt} & \textbf{System Prompt} & \textbf{User Prompt} \\
\hline
Text-only &
\texttt{Task: Deconstruct the following sentence into its underlying logical structure. Describe the meaning by identifying core events, the entities involved, and how they relate to one another.\newline
Example:\newline
Input Sentence: "The NBA season of 1975 -- 76 was the 30th season of the National Basketball Association ." \newline
Output: This refers to a season of the National Basketball Association. It was the 30th season of the league. The season took place during the 1975-76 time period.}
&
\texttt{Sentence: ``Captain'' was hulked in 1739 , and eventually broken up in 1762 .}
\\
\hline
Text + AMR &
\texttt{Task: Deconstruct the following sentence into its underlying logical structure. Describe the meaning by identifying core events, the entities involved, and how they relate to one another. You may use the provided AMR (Abstract Meaning Representation) as a supplement to guide you through generation.\newline
Example:\newline
Input Sentence: "The NBA season of 1975 -- 76 was the 30th season of the National Basketball Association ."\newline
Input AMR:\newline
(s / season\textasciitilde2\newline
\phantom{xxxxx}:ord (o / ordinal-entity\textasciitilde10\newline
\phantom{xxxxxxxxxx}:value 30\textasciitilde9)\newline
\phantom{xxxxx}:poss (l / league\textasciitilde13\newline
\phantom{xxxxxxxxxx}:name (n / name\textasciitilde13\newline
\phantom{xxxxxxxxxxxxxxx}:op1 "National"\textasciitilde13\newline
\phantom{xxxxxxxxxxxxxxx}:op2 "Basketball"\textasciitilde14\newline
\phantom{xxxxxxxxxxxxxxx}:op3 "Association"\textasciitilde15))\newline
\phantom{xxxxx}:time (d3 / date-interval\textasciitilde4\newline
\phantom{xxxxxxxxxx}:op1 (d / date-entity\textasciitilde4\newline
\phantom{xxxxxxxxxxxxxxx}:year 1975\textasciitilde4)\newline
\phantom{xxxxxxxxxx}:op2 (d2 / date-entity\textasciitilde6\newline
\phantom{xxxxxxxxxxxxxxx}:year 76\textasciitilde6)))\newline
Output: This refers to a season of the National Basketball Association. It was the 30th season of the league. The season took place during the 1975-76 time period.}
&
\texttt{Sentence: ``Captain'' was hulked in 1739 , and eventually broken up in 1762 .\newline
AMR:\newline 
(a / and\newline
\phantom{xxxxx}:op1 (h / hulk-01\newline
\phantom{xxxxxxxxxx}:ARG1 (s / ship\newline
\phantom{xxxxxxxxxxxxxxx}:name (n / name\newline
\phantom{xxxxxxxxxxxxxxxxxx}:op1 "Captain"))\newline
\phantom{xxxxxxxxxx}:time (d / date-entity\newline
\phantom{xxxxxxxxxxxxxxx}:year 1739))\newline
\phantom{xxxxx}:op2 (b / break-up-08\newline
\phantom{xxxxxxxxxx}:ARG1 s\newline
\phantom{xxxxxxxxxx}:time (e / eventual)\newline
\phantom{xxxxxxxxxx}:time (d2 / date-entity\newline
\phantom{xxxxxxxxxxxxxxx}:year 1762)))}
\\
\hline
Text + AMR nodes &
\texttt{Task: Deconstruct the following sentence into its underlying logical structure. Describe the meaning by identifying core events, the entities involved, and how they relate to one another. You may use the supplement provided if you find it helpful.\newline
Example:\newline
Input Sentence: "The NBA season of 1975 -- 76 was the 30th season of the National Basketball Association ." \newline
Supplement: season, ordinal-entity, 30, league, name, "National", "Basketball", "Association", date-interval, date-entity, 1975, date-entity, 76\newline
Output: This refers to a season of the National Basketball Association. It was the 30th season of the league. The season took place during the 1975-76 time period.}
&
\texttt{Sentence: ``Captain'' was hulked in 1739 , and eventually broken up in 1762 .\newline
Supplement: and, hulk-01, ship, name, "Captain", date-entity, 1739, break-up-08, s, eventual, date-entity, 1762}
\\
\hline
\end{tabularx}
}
\caption{Prompt types and their corresponding system prompt}
\label{tab:ppl-prompt}
\end{table*}

\subsection{AMR-NLD} \label{amr-nld-prompt}
Figure~\ref{lst:system-prompt} presents the full system prompt used for generating AMR-NLD.

\onecolumn
\begin{tcolorbox}[
    title=\textbf{Full system prompt for AMR-to-NLD generation},
    fonttitle=\small,
    colback=gray!5,
    colframe=black!60,
    left=6pt, right=6pt, top=4pt, bottom=4pt,
    breakable
]
\small

You are an expert at reading AMR graphs and describing them in plain natural language. Given an AMR graph, decompose it into its core relational branches (such as \texttt{:ord}, \texttt{:poss}, \texttt{:time}, \texttt{:ARG0}, \texttt{:ARG1}, \texttt{:location}, etc.), map each branch to an individual natural language description, then aggregate all descriptions into a final coherent AMR-NLD (Natural Language Description).

\medskip
\textbf{Rules:}
\begin{itemize}[leftmargin=1.5em, itemsep=0pt, topsep=2pt]
    \item No technical terms like ARG0, ARG1, op1, variable names, etc.
    \item Decompose the AMR into its relational branches first, then map each to a sentence.
    \item Each sentence should capture one relationship or branch from the AMR.
    \item Aggregate the individual descriptions into a coherent, flowing final output.
    \item Use natural pronouns and nouns to handle repeated entities.
    \item If an argument is missing, skip it; do not mention it is absent.
    \item Keep sentences simple and direct.
    \item Ignore all index markers like \texttt{\textasciitilde1}, \texttt{\textasciitilde2}, \texttt{\textasciitilde3} in the AMR.
    \item Output ONLY the final aggregated sentences, no preamble, no intro like ``Here are the sentences...'' or any other meta-commentary.
\end{itemize}

\medskip
\noindent\rule{\linewidth}{0.3pt}

\textbf{Example 1:}\\[2pt]
\textbf{AMR:}
\begin{tcolorbox}[
    colback=white, colframe=black!30,
    left=4pt, right=4pt, top=2pt, bottom=2pt,
    boxsep=0pt, arc=2pt
]
\begin{flushleft}
\fontsize{6}{7.5}\ttfamily
(g / give-01\\
\hspace*{1em}:ARG0 (p / person :name (n / name :op1 "John"))\\
\hspace*{1em}:ARG1 (b / book)\\
\hspace*{1em}:ARG2 (p2 / person :name (n2 / name :op1 "Mary")))
\end{flushleft}
\end{tcolorbox}
\textbf{Branch decomposition:}
\begin{itemize}[leftmargin=1.5em, itemsep=0pt, topsep=2pt]
    \item \texttt{:ARG0} $\rightarrow$ John is the giver
    \item \texttt{:ARG1} $\rightarrow$ a book is what was given
    \item \texttt{:ARG2} $\rightarrow$ Mary is the recipient
\end{itemize}
\textbf{Output:}\\
John gave a book. The recipient of the book was Mary.

\medskip
\noindent\rule{\linewidth}{0.3pt}

\textbf{Example 2:}\\[2pt]
\textbf{AMR:}
\begin{tcolorbox}[
    colback=white, colframe=black!30,
    left=4pt, right=4pt, top=2pt, bottom=2pt,
    boxsep=0pt, arc=2pt
]
\begin{flushleft}
\fontsize{6}{7.5}\ttfamily
(s / season\\
\hspace*{1em}:ord (o / ordinal-entity :value 30)\\
\hspace*{1em}:poss (l / league :name (n / name :op1 "National" :op2 "Basketball" :op3 "Association"))\\
\hspace*{1em}:time (d / date-entity :year 1975 :year2 1976))
\end{flushleft}
\end{tcolorbox}
\textbf{Branch decomposition:}
\begin{itemize}[leftmargin=1.5em, itemsep=0pt, topsep=2pt]
    \item \texttt{:poss} $\rightarrow$ belongs to the National Basketball Association
    \item \texttt{:ord} $\rightarrow$ it was the 30th season
    \item \texttt{:time} $\rightarrow$ it took place in the 1975--76 period
\end{itemize}
\textbf{Output:}\\
This refers to a season of the National Basketball Association. It was the 30th season of the league. The season took place during the 1975--76 time period.

\medskip
\noindent\rule{\linewidth}{0.3pt}

\textbf{Example 3:}\\[2pt]
\textbf{AMR:}
\begin{tcolorbox}[
    colback=white, colframe=black!30,
    left=4pt, right=4pt, top=2pt, bottom=2pt,
    boxsep=0pt, arc=2pt
]
\begin{flushleft}
\fontsize{6}{7.5}\ttfamily
(s / sell-off-04\\
\hspace*{1em}:ARG0 (g3 / group\\
\hspace*{2em}:consist-of (p / person :ARG0-of (h / hack-04))\\
\hspace*{2em}:mod (c3 / criminal-organization :name (n / name :op1 "Shadow" :op2 "Brokers")))\\
\hspace*{1em}:ARG1 (t / tool\\
\hspace*{2em}:purpose (s2 / spy-01 :mod (c5 / cyber))\\
\hspace*{2em}:ARG0-of (b / belong-01\\
\hspace*{3em}:ARG1 (g2 / government-organization\\
\hspace*{4em}:ARG0-of (g / govern-01 :ARG1 (c2 / country :name (n2 / name :op1 "U.S."))))\\
\hspace*{3em}:ARG1-of (c / claim-01 :ARG0 g3)))\\
\hspace*{1em}:manner (a / auction-02 :mod (o / online))\\
\hspace*{1em}:time (c4 / current))
\end{flushleft}
\end{tcolorbox}
\textbf{Branch decomposition:}
\begin{itemize}[leftmargin=1.5em, itemsep=0pt, topsep=2pt]
    \item \texttt{:ARG0} $\rightarrow$ a criminal organization called the Shadow Brokers, consisting of hackers
    \item \texttt{:ARG1} $\rightarrow$ cyberspy tools claimed to belong to the U.S. government
    \item \texttt{:manner} $\rightarrow$ conducted via online auction
    \item \texttt{:time} $\rightarrow$ currently ongoing
\end{itemize}
\textbf{Output:}\\
A criminal organization called the Shadow Brokers conducted a sell-off. The group consisted of hackers. They sold cyberspy tools. The Shadow Brokers claimed these tools belonged to the U.S. government. The auction was conducted online. This sell-off is currently ongoing.

\medskip
\noindent\rule{\linewidth}{0.3pt}

\textbf{Example 4:}\\[2pt]
\textbf{AMR:}
\begin{tcolorbox}[
    colback=white, colframe=black!30,
    left=4pt, right=4pt, top=2pt, bottom=2pt,
    boxsep=0pt, arc=2pt
]
\begin{flushleft}
\fontsize{6}{7.5}\ttfamily
(e / exterminate-01\\
\hspace*{1em}:ARG0 (p / person :name (n / name :op1 "Saddam" :op2 "Hussein"))\\
\hspace*{1em}:ARG1 (o / person :ARG0-of (o2 / oppose-01))\\
\hspace*{1em}:location (c / country :name (n2 / name :op1 "Iraq")))
\end{flushleft}
\end{tcolorbox}
\textbf{Branch decomposition:}
\begin{itemize}[leftmargin=1.5em, itemsep=0pt, topsep=2pt]
    \item \texttt{:ARG0} $\rightarrow$ Saddam Hussein is the one who exterminated
    \item \texttt{:ARG1} $\rightarrow$ the targets were people who opposed him
    \item \texttt{:location} $\rightarrow$ this took place in Iraq
\end{itemize}
\textbf{Output:}\\
Saddam Hussein carried out an extermination. The targets of this extermination were people who opposed him. This took place in Iraq.

\medskip
\noindent\rule{\linewidth}{0.3pt}

\textbf{Example 5:}\\[2pt]
\textbf{AMR:}
\begin{tcolorbox}[
    colback=white, colframe=black!30,
    left=4pt, right=4pt, top=2pt, bottom=2pt,
    boxsep=0pt, arc=2pt
]
\begin{flushleft}
\fontsize{6}{7.5}\ttfamily
(d / decry-01\\
\hspace*{1em}:ARG0 (p2 / person :name (n / name :op1 "Trump"))\\
\hspace*{1em}:ARG1 (a / and\\
\hspace*{2em}:op1 (p4 / protect-01\\
\hspace*{3em}:ARG0 (c / country :name (n2 / name :op1 "US"))\\
\hspace*{3em}:ARG1 (c2 / country :name (n3 / name :op1 "Saudi" :op2 "Arabia")))\\
\hspace*{2em}:op2 (r / reimburse-01\\
\hspace*{3em}:ARG1 (p / penny\\
\hspace*{4em}:ARG3-of (s2 / spend-01 :ARG0 (w / we)\\
\hspace*{5em}:time (s / sit-01 :ARG1 c2\\
\hspace*{6em}:ARG2 (m2 / multiple\\
\hspace*{7em}:op1 (m / monetary-quantity :quant 1000000000000 :unit (d2 / dollar))))))\\
\hspace*{3em}:mod (e / every))\\
\hspace*{2em}:ARG2 c\\
\hspace*{2em}:manner (p3 / proper)\\
\hspace*{2em}:polarity -))
\end{flushleft}
\end{tcolorbox}
\textbf{Branch decomposition:}
\begin{itemize}[leftmargin=1.5em, itemsep=0pt, topsep=2pt]
    \item \texttt{:ARG0} $\rightarrow$ Trump is the one decrying
    \item \texttt{:ARG1 :op1} $\rightarrow$ the US protecting Saudi Arabia
    \item \texttt{:ARG1 :op2} $\rightarrow$ Saudi Arabia not properly reimbursing the US for every penny spent
    \item (context) $\rightarrow$ Saudi Arabia sits on over a trillion dollars
\end{itemize}
\textbf{Output:}\\
Trump decried two things. First, he decried the fact that the US was protecting Saudi Arabia. Second, he decried that Saudi Arabia had not properly reimbursed the US for every penny that the US spent on protecting Saudi Arabia. This was particularly troubling given that Saudi Arabia was sitting on over a trillion dollars.

\end{tcolorbox}
\captionof{figure}{Full system prompt for AMR-to-NLD generation}
\label{lst:system-prompt}

\clearpage
\twocolumn

\end{document}